\documentclass{article}

\usepackage[preprint]{preprint}

\usepackage[utf8]{inputenc} 
\usepackage[T1]{fontenc}    
\usepackage{hyperref}       
\usepackage{url}            
\usepackage{booktabs}       
\usepackage{amsfonts}       
\usepackage{nicefrac}       
\usepackage{microtype}      
\usepackage{xcolor}         
\usepackage{graphicx}
\usepackage{float}
\usepackage{amsmath,amssymb,amsthm}
\usepackage{makecell}
\usepackage{subcaption}
\usepackage{placeins}
\usepackage{enumitem}
\usepackage[colorinlistoftodos,prependcaption,textsize=small]{todonotes}

\usepackage{tikz}
\usetikzlibrary{arrows.meta,calc,positioning}

\newtheorem{definition}{Definition}

\newcommand{\E}{\mathbb E}
\newcommand{\Cov}{\operatorname{Cov}}
\newcommand{\Var}{\operatorname{Var}}
\newcommand{\cY}{\mathcal Y}
\newcommand{\cQ}{\mathcal Q}
\newcommand{\cK}{\mathcal K}

\title{Elicitation without Backpropagation: Steering Model Behavior by Optimizing the Latent Posterior}

\author{%
  Garrett Baker \\
  Timaeus \\
  \texttt{garrett@timaeus.co} \\
  \And
  Vinayak Pathak\thanks{Work done while at Timaeus.} \\
  CORAL \\
  \texttt{path.vinayak@gmail.com} \\
  \And
  Daniel Murfet \\
  Timaeus \\
  \texttt{daniel@timaeus.co} \\
  \And
  Susan Wei\thanks{Corresponding author.} \\
  Monash University \\
  \texttt{susan.wei@monash.edu} \\
}

\begin{document}
\raggedbottom

\maketitle

\begin{abstract}
  In the \emph{latent posterior model} of transformer behavior, the next-token distribution arises from a posterior over latent predictive models conditioned on the context, mixed to generate continuations. We exploit this model in settings where it is exact, namely Bayes-filtered transformers (BFTs) meta-learned on sequences from a hierarchical prior, to introduce \textbf{Posterior Prefix Tuning (PPT)}, a new method for \emph{eliciting} behavior from a transformer: given a utility function on continuations, find a prompt under which the transformer generates continuations of high expected utility. For a BFT, the elicitation objective factors through the latent posterior, and the gradient of this objective can be estimated from samples of the prior alone. PPT optimizes the parameters of a distribution over hard prompts: it draws prior samples once from the BFT via predictive Monte Carlo (PMC), then estimates the gradient by importance sampling against them. The optimization performs no transformer forward passes and no backpropagation through the transformer, and the prior samples are utility-independent, so a single set of samples drives elicitation against any number of utilities at negligible marginal cost. We validate PPT on Beta--Bernoulli and reinforced urn BFTs across three utility families (reverse cross-entropy, frequency matching, Dyck validity).
\end{abstract}

Code is available at \url{https://github.com/timaeus-research/elicitation}.

\section{Introduction}
\label{sec:introduction}

Transformers trained on distributions with diverse tasks exhibit \emph{in-context learning} (ICL), the ability to learn from context alone \citep{brown2020language}. One important model of this behavior, the \textbf{latent posterior model}, holds that the transformer maintains a posterior over latent predictive models, updated by the tokens it has seen, and uses this posterior to generate continuations \citep{xie2022explanationincontextlearningimplicit, panwar2024incontextlearningbayesianprism, marks2026persona}. The latent posterior model has been verified as a good approximation in synthetic settings such as in-context linear regression \citep{garg2022what, akyurek2023learningalgorithmincontextlearning}, and has the potential to confer new ideas for understanding and controlling transformer behavior more broadly.

\textbf{In this paper we explore this potential, motivated by the elicitation problem.} We currently lack a deep scientific understanding of why transformers, post-trained to be effective assistants, behave (and misbehave) in the ways that they do. These systems undergo testing before deployment, which can be formalized as the evaluation of utility functions $U$ on the model's outputs across a distribution of prompts. However, the space of possible prompts is too vast to search exhaustively, so a prompt that elicits undesirable behavior can slip through testing. This leads to the problem of elicitation \citep{irving2025eliciting}: given a specification of undesirable behavior (formalized as a utility function $U$), which prompt maximizes the expected utility of the transformer's continuations? This is closely related to the problem of prompt and prefix tuning \citep{li2021prefix,lester2021power}.

A promising avenue for investigating the elicitation problem is to factor it through the latent posterior. Rather than searching directly over the combinatorial space of prompts, we ask: is there a prompt conditioned on which the latent posterior concentrates on predictive models with a high probability of generating undesirable behavior? We study this question in \textbf{Bayes-filtered transformers} (BFTs) \citep{fortini2026uncertainty}: transformers meta-learned on sequences from a two-stage hierarchical process, in which a latent task is first sampled from a prior and the sequence is then generated from the task's likelihood. For such sequences, the autoregressive log-loss is minimized at every position by the Bayesian posterior predictive, determined jointly by the prior over latent tasks and the likelihood by which each task generates the sequence \citep{ortegaMetalearningSequentialStrategies2019a}. We call a BFT that attains this minimum exactly the \emph{idealized BFT}; it realizes the latent posterior model by construction, and a suitably trained BFT approximately admits the same factorization. For a BFT, elicitation can therefore be carried out entirely in latent-posterior space. We call the resulting method \textbf{Posterior Prefix Tuning} (PPT).

\begin{figure}[t]
  \centering
  \definecolor{rust}{RGB}{195,75,50}
  \begin{tikzpicture}
    \node[anchor=south west, inner sep=0] (img1) at (0, 0)
    {\includegraphics[width=0.32\textwidth]{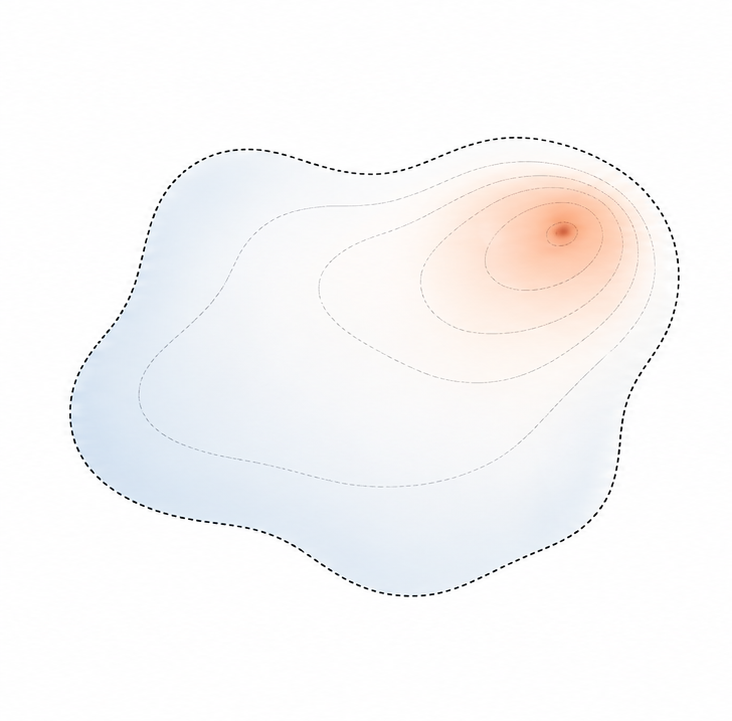}};
    \node[anchor=south west, inner sep=0] (img2) at (0.34\textwidth, 0)
    {\includegraphics[width=0.32\textwidth]{figures/fig1_small.png}};
    \node[anchor=south west, inner sep=0] (img3) at (0.68\textwidth, 0)
    {\includegraphics[width=0.32\textwidth]{figures/fig1_small.png}};

    \def\dotlist{%
      0.20/0.27, 0.35/0.25, 0.55/0.22, 0.75/0.30,
      0.13/0.45, 0.30/0.40, 0.42/0.32, 0.48/0.42, 0.65/0.43, 0.83/0.40,
      0.18/0.58, 0.34/0.56, 0.52/0.56, 0.68/0.55, 0.85/0.55,
      0.22/0.72, 0.40/0.70, 0.55/0.72, 0.70/0.65, 0.78/0.78,
      0.25/0.65, 0.30/0.80, 0.48/0.82, 0.62/0.78,
      0.76/0.68, 0.83/0.72%
    }

    \begin{scope}[shift={(img2.south west)},
        x={($(img2.south east)-(img2.south west)$)},
        y={($(img2.north west)-(img2.south west)$)}]
      \foreach \x/\y in \dotlist {
        \pgfmathsetmacro{\xc}{0.5 + 0.85*(\x-0.5)}
        \pgfmathsetmacro{\yc}{0.5 + 0.75*(\y-0.5)}
        \fill[black!60] (\xc, \yc) circle (1.4pt);
      }
    \end{scope}

    \begin{scope}[shift={(img3.south west)},
        x={($(img3.south east)-(img3.south west)$)},
        y={($(img3.north west)-(img3.south west)$)}]
      \foreach \x/\y in \dotlist {
        \pgfmathsetmacro{\xc}{0.5 + 0.85*(\x-0.5)}
        \pgfmathsetmacro{\yc}{0.5 + 0.75*(\y-0.5)}
        \pgfmathsetmacro{\dpeak}{sqrt((\xc-0.74)*(\xc-0.74) + (\yc-0.70)*(\yc-0.70))}
        \pgfmathsetmacro{\rsize}{3.0 * exp(-2.5*\dpeak)}
        \fill[rust] (\xc, \yc) circle (\rsize pt);
      }
    \end{scope}

    \begin{scope}[x={(img3.south east)}, y={(img1.north west)}]
      \node[text=black, font=\footnotesize\bfseries] at (0.18, 0.95) {The latent landscape};
      \node[text=black, font=\footnotesize\bfseries] at (0.52, 0.95) {Latent prior samples};
      \node[text=black, font=\footnotesize] at (0.52, 0.87) {(utility independent)};
      \node[text=black, font=\footnotesize\bfseries] at (0.85, 0.95) {Latent posterior samples};
      \node[text=black, font=\footnotesize] at (0.85, 0.87) {(obtained by tilting)};

      \node[font=\footnotesize] (cK) at (0.08, 0.06) {$\cK$};
      \draw[->,black] (cK) -- (0.08, 0.40);

      \node[font=\footnotesize] (highutil) at (0.25, 0.06) {High $\mu(\tilde{k})$};
      \draw[->,black] (highutil) -- (0.25, 0.62);

      \node[font=\footnotesize] (callpi) at (0.50, 0.06) {Roll out $\pi_\theta$ to sample its prior};
      \draw[->,black] (callpi) -- (0.50, 0.42);

      \node[font=\footnotesize] (findz) at (0.84, 0.06) {Optimize $\kappa$ so the latent};
      \node[font=\footnotesize] at (0.84, 0.00) {posterior weights high utility $\tilde{k}$};
      \draw[->,black] (findz) -- (0.91, 0.59);
    \end{scope}
  \end{tikzpicture}

  \vspace{1em}

  \caption{
    \textbf{Steering the latent posterior to elicit behavior.} Given a utility $U$ on continuations, let $\mu(\tilde k)$ be its expectation under a latent $\tilde k$. To elicit high-utility behavior from the transformer $\pi_\theta$, we seek prompts $z_{1:m}$ that steer the latent posterior $\Pi(d\tilde k \mid z_{1:m})$ over $\cK$ toward latents $\tilde{k}$ with high $\mu(\tilde{k})$. \emph{Left}: the latent space $\cK$. \emph{Middle}: Predictive Monte Carlo (PMC) samples once from the latent prior the BFT implies; this is the only time we call $\pi_\theta$, and it does not use $U$. \emph{Right}: a continuous surrogate $\kappa$ for $z_{1:m}$ and a tilt $W(\tilde{k},\kappa)$ define a tilted posterior $\Pi_{\mathrm{tilt}}(d\tilde k;\, \kappa)$, sampled by importance reweighting of the fixed prior samples; optimizing $\kappa$ concentrates this posterior on latents with high $\mu(\tilde k)$. Snapping the optimum yields a hard prompt $z_{1:m}$.
  }
  \label{fig:latent-posterior}
\end{figure}

Figure~\ref{fig:latent-posterior} sketches PPT. The method has two properties:
\begin{itemize}[leftmargin=1.5em]
  \item \textbf{No backpropagation through the transformer.} PPT computes the gradient of its objective entirely in latent-posterior space, so optimization never backpropagates through the transformer and makes zero transformer calls per step.

  \item \textbf{Utility amortization.} The bulk of PPT's computation is utility-independent: characterizing the BFT's beliefs (the prior samples in the middle panel of Figure~\ref{fig:latent-posterior}) is a one-time cost, after which any number of utility functions can be optimized against at negligible marginal cost.


\end{itemize}

We instantiate this approach with two BFTs: one meta-learned on an exchangeable process, whose latent task is a probability mass function (pmf), and one on a 1-Markov exchangeable process, whose latent task is a transition matrix. Both latent tasks are finite-dimensional, so we optimize over a prompt distribution of the same form and recover a hard prompt from the optimum.

We find that PPT and its Rao--Blackwellized variant PPT-RB are effective across both BFTs, three utility families (reverse cross-entropy, frequency matching, Dyck validity), and prompt lengths $m \in \{6, 50\}$. The comparison with Greedy Coordinate Gradient (GCG) \citep{zou2023universal}, the standard hard-prompt baseline, is mixed: PPT methods substantially outperform GCG on the reinforced urn at $m=6$ (PPT-RB reaches the Dyck optimum on every seed) and on reverse cross-entropy on the reinforced urn at $m=50$, while GCG is competitive or stronger on the Beta--Bernoulli BFT and on Dyck validity for the reinforced urn at $m=50$. We note that GCG backpropagates through the transformer at every step, unlike PPT and PPT-RB.

\textbf{Outline.} The remainder of the paper is organized as follows. Section~\ref{sec:overview} establishes the formal setup and connects the elicitation problem to the latent-posterior factorization \eqref{eq:icb-general}. Section~\ref{sec:ppt} develops PPT and derives the gradient estimator. The experimental results follow in the remainder.

\section{Overview}
\label{sec:overview}

\subsection{Formalizing the elicitation problem}
\label{sec:formal_elicitation}
Let $\cY$ be a finite vocabulary, $\Delta(\cY)$ the simplex of pmfs on $\cY$, and $\cY^* := \bigcup_{n\ge 0}\cY^n$. Let $\pi_\theta:\cY^*\to\Delta(\cY)$, $y_{1:n}\mapsto\pi_\theta(\cdot\mid y_{1:n})$, be a transformer sequence model, with empty history $y_{1:0}:=\varnothing$. In the prefix-tuning setting, the prompt $z_{1:m}\in\cY^m$ is fed to $\pi_\theta$ as a prefix. Fix a continuation horizon $N\geq 1$ and let $Y_{1:N}\in\cY^N$ denote the length-$N$ continuation generated autoregressively from $\pi_\theta$ given $z_{1:m}$, i.e.\ $Y_t\sim\pi_\theta(\cdot\mid z_{1:m},Y_{1:t-1})$ for $t=1,\dots,N$; abusing notation, we write $\pi_\theta(\cdot\mid z_{1:m})$ for the resulting law of $Y_{1:N}$, extending the one-step predictor $\pi_\theta$ to the continuation. We formalize undesirable behavior via a utility function $U:\cY^N\to\mathbb{R}$ evaluated on this continuation. Given a hard prompt $z_{1:m}\in\cY^m$, the \emph{elicitation objective} is
\begin{equation}
  J(z_{1:m}) := \E_{Y_{1:N}\sim\pi_\theta(\cdot\mid z_{1:m})}\!\big[U(Y_{1:N})\big].
  \label{eq:J}
\end{equation}

\begin{definition}[Elicitation problem]
  Given a sequence model $\pi_\theta$, a utility function $U$, and a set of allowed prompts $\mathcal{Z} \subseteq \cY^m$, the elicitation problem is to find $z_{1:m} \in \mathcal{Z}$ maximizing $J(z_{1:m})$.
\end{definition}

\subsection{Bayes-filtered transformers and the latent-posterior factorization}
\label{sec:overview_icb}

\textbf{Bayes-filtered transformers.} A Bayes-filtered transformer (BFT) is one meta-learned on sequences from a two-stage hierarchical process: a latent task $\tilde k$ is drawn from a prior $\Pi_0$ on a class $\cK := \{\tilde k:\cY^*\to\Delta(\cY)\}$ of kernels,\footnote{Throughout, \emph{kernel} is used in its probabilistic sense: a map taking a history $h\in\cY^*$ to a probability distribution $\tilde k(\cdot\mid h)\in\Delta(\cY)$ over the next token, equivalently a conditional distribution of the next symbol given the past. This is the notion of a Markov (also transition or stochastic) kernel; it specializes to a pmf when there is no history dependence ($k=0$, Section~\ref{sec:ppt_k0}) and to a transition matrix in the $1$-Markov case (Section~\ref{sec:ppt_kmarkov}). It is unrelated to the kernel of kernel methods.} and a sequence $y_{1:n}$ is then generated from $\tilde k$, i.e.\ $y_i\sim\tilde k(\cdot\mid y_{1:i-1})$. The predictor $\pi_\theta:\cY^*\to\Delta(\cY)$ is trained to minimize the \emph{population risk}, the expected autoregressive log-loss under this process,
\begin{equation}
  R(\pi_\theta) := \E_{\tilde k\sim\Pi_0}\;\E_{y_{1:n}\sim\tilde k}\Big[\,\textstyle\sum_{i=1}^n -\log\pi_\theta(y_i\mid y_{1:i-1})\,\Big].
  \label{eq:population_risk}
\end{equation}
Minimized over all predictors, \eqref{eq:population_risk} is solved at every position by the Bayesian posterior predictive distribution (PPD), determined jointly by the prior $\Pi_0$ and the likelihood by which each task generates the sequence, given the observed prefix \citep{ortegaMetalearningSequentialStrategies2019a}; the training targets are \emph{Bayes-filtered} in this sense, and the term BFT follows \citet{fortini2026uncertainty}.

\textbf{Meta-learning a BFT.} The population risk \eqref{eq:population_risk} is not directly accessible, so meta-learning minimizes its empirical counterpart (empirical risk minimization): drawing a training set $\{y^{(j)}_{1:n}\}_{j=1}^D$ from the process, $\tilde k^{(j)}\sim\Pi_0$ then $y^{(j)}_{1:n}\sim\tilde k^{(j)}$, we minimize
\begin{equation}
  \widehat R(\pi_\theta) := \frac{1}{D}\sum_{j=1}^D \sum_{i=1}^n -\log\pi_\theta(y^{(j)}_i\mid y^{(j)}_{1:i-1}).
  \label{eq:empirical_risk}
\end{equation}
The loss itself is standard for transformer pretraining; what distinguishes BFT training is this data.

In the limit of infinite data, infinite capacity, and perfect optimization, the trained transformer attains the population minimum of \eqref{eq:population_risk}, realizing the PPD exactly; we call this limit the \emph{idealized BFT}. With finite data, parameters, and optimization, it only approximates the idealized BFT, and the quality of this approximation has been studied empirically across a range of settings \citep{mikulikMetatrainedAgentsImplement2020, geneweinMemoryBasedMetaLearningNonStationary2023, grau-moyaLearningUniversalPredictors2024}.

\textbf{The latent-posterior factorization.} Concretely, the minimizer of \eqref{eq:population_risk} is a Bayes mixture over the latent kernel: when the trained BFT realizes the PPD, as the idealized BFT does, its one-step predictive distribution is
\begin{equation}
  \pi_\theta(y \mid y_{1:n})
  =
  \int_{\cK} \tilde k(y \mid y_{1:n}) \,\Pi(d\tilde k \mid y_{1:n}), \qquad y\in\cY,
  \label{eq:icb-general}
\end{equation}
where $\Pi(\cdot \mid y_{1:n})$ is the posterior under $\Pi_0$ given the prefix:
\[
  \Pi(d\tilde k \mid y_{1:n}) \,\propto\, \Big[\textstyle\prod_{i=1}^n \tilde k(y_i \mid y_{1:i-1})\Big]\,\Pi_0(d\tilde k).
\]
This factorization holds exactly for the idealized BFT; a trained $\pi_\theta$ inherits it asymptotically, for large prefix lengths $n$, when the approximation error is well-behaved.

\textbf{Factoring $J$ through the latent posterior.} Let us extend $\tilde k$ from one-step predictions to length-$N$ continuations via the autoregressive product
$
  \tilde k(y_{1:N}\mid z_{1:m}) \,:=\, \prod_{t=1}^N \tilde k(y_t\mid z_{1:m},\, y_{1:t-1}),
$
i.e., the joint probability that the next $N$ tokens equal $y_{1:N}$ when each is sampled from $\tilde k$ given the prefix accumulated so far. Set
\[
  \mu(\tilde k;\, z_{1:m}) \,:=\, \sum_{y_{1:N}\in\cY^N} U(y_{1:N})\, \tilde k(y_{1:N}\mid z_{1:m}),
\]
the expected $U$ over continuations sampled from $\tilde k$ given $z_{1:m}$.
Under \eqref{eq:icb-general}, the tower property yields
\begin{equation}
  J(z_{1:m})
  =
  \int_{\cK} \mu(\tilde k;\, z_{1:m})\,\Pi(d\tilde k \mid z_{1:m}).
  \label{eq:tower_identity}
\end{equation}

\section{Posterior Prefix Tuning (PPT)}
\label{sec:ppt}
The factorization \eqref{eq:tower_identity} expresses $J(z_{1:m})$ as a posterior expectation, but optimizing it directly remains a discrete search over $z_{1:m}\in\cY^m$. Existing hard-prompt methods such as greedy coordinate gradient (GCG) \citep{zou2023universal} tackle this discreteness by backpropagating through the transformer's embedding layer for gradient signal in $z_{1:m}$, requiring per-step transformer calls and gradients.

Our proposed method, \textbf{Posterior Prefix Tuning (PPT)}, instead considers a kernel $\kappa\in\cK$ in the same class as the latent task and optimizes a surrogate $J_{\mathrm{tilt}}$ in $\kappa$. The construction has four pieces. The \emph{tilt factor} $W(\tilde k;\,\kappa)$ is the probability that a latent task $\tilde k$ assigns to a prompt drawn from $\kappa$, measuring how compatible $\tilde k$ is with $\kappa$. Reweighting the prior $\Pi_0$ by this factor gives the \emph{tilted posterior} $\Pi_{\mathrm{tilt}}(\cdot;\,\kappa)\propto W(\tilde k;\,\kappa)\,\Pi_0$, which favors the tasks most compatible with $\kappa$. Averaging a task's utility over prompts drawn from $\kappa$ gives the \emph{marginalized utility} $\bar\mu(\tilde k;\,\kappa)$, and integrating it against the tilted posterior gives the \emph{surrogate objective} $J_{\mathrm{tilt}}(\kappa)$, the tower identity \eqref{eq:tower_identity} with $\Pi(\cdot\mid z_{1:m})$ replaced by $\Pi_{\mathrm{tilt}}(\cdot;\,\kappa)$ and $\mu$ by $\bar\mu$. Table~\ref{tab:tilt-objects} gives these four objects; their derivations are in Appendix~\ref{app:ppt_details}.

\begin{table}[h]
  \centering
  \caption{The four objects on which PPT is built.}
  \label{tab:tilt-objects}
  \begin{tabular}{@{}l l@{}}
    \toprule
    Tilt factor          & $W(\tilde k;\, \kappa) := \E_{z_{1:m}\sim\kappa}\!\big[\textstyle\prod_{i=1}^m \tilde k(z_i\mid z_{1:i-1})\big]$           \\[4pt]
    Tilted posterior     & $\Pi_{\mathrm{tilt}}(d\tilde k;\, \kappa) \propto W(\tilde k;\, \kappa)\,\Pi_0(d\tilde k)$                            \\[4pt]
    Marginalized utility & $\bar\mu(\tilde k;\, \kappa) := \E_{z_{1:m}\sim\kappa}\!\big[\mu(\tilde k;\, z_{1:m})\big]$                            \\[4pt]
    Surrogate objective  & $J_{\mathrm{tilt}}(\kappa) := \int_\cK \bar\mu(\tilde k;\, \kappa)\,\Pi_{\mathrm{tilt}}(d\tilde k;\, \kappa)$           \\
    \bottomrule
  \end{tabular}
\end{table}

To optimize over $\kappa$ we need a finite-dimensional parameterization. We therefore work with $k$-Markov kernels: a kernel $\tilde k\in\cK$ is \emph{$k$-Markov} if $\tilde k(\cdot\mid h)$ depends on $h$ only through its length-$k$ suffix $\mathrm{suf}_k(h)$, the last $k$ symbols of $h$ (we write $\cK_k\subseteq\cK$ for this subset; formal background in Appendix~\ref{app:markov-exch}). A $k$-Markov kernel is finite-dimensional: a pmf in $\Delta(\cY)$ when $k=0$, and a transition table in $\cQ_k$ when $k\geq 1$. Taking $\Pi_0$ supported on $\cK_k$, both the latent task and the prompt kernel $\kappa$ are of this form. Sections~\ref{sec:ppt_k0} and~\ref{sec:ppt_kmarkov} instantiate these for $k=0$ and $k\geq 1$.

\begin{figure}[!h]
  \makebox[\textwidth][c]{%
    \begin{subfigure}[t]{0.55\columnwidth}
      \centering
      \resizebox{\linewidth}{!}{%
        \begin{tikzpicture}[
            xscale=1.0, yscale=0.85,
            rollout/.style={font=\scriptsize},
            suffix/.style={text=black, fill=blue!15, rounded corners, inner sep=2pt, minimum width=0.7cm, align=center},
            arrow/.style={->, thick, >=Stealth},
            header/.style={align=center, font=\tiny\sffamily, anchor=south}
          ]

          \node[header] (h2) at (1.4, 0.4) {Autoregressive sampling via $\pi_\theta$ \\ $y^{(l)}_t \sim \pi_\theta(\cdot \mid y^{(l)}_{1:t-1})$};
          \node[header] (h3) at (4.1, 0.4) {Approximate\\ prior sample};

          \node[font=\sffamily\tiny, anchor=east] at (-0.4, 0) {Rollout 1};
          \node[rollout, suffix] (s1_1) at (0, 0) {$y^{(1)}_1$};
          \node (s1_d) at (0.9, 0) {$\dots$};
          \node[rollout, suffix] (s1_N) at (1.9, 0) {$y^{(1)}_R$};
          \node[font=\scriptsize] (theta1) at (4.1, 0) {$\tilde p^{(1)}$ or $\tilde Q^{(1)}$};
          \draw[arrow] (s1_N) -- (theta1) node[midway, above=0.01cm, font=\tiny\sffamily] {count};

          \node[font=\sffamily\tiny, anchor=east] at (-0.4, -0.9) {Rollout 2};
          \node[rollout, suffix] (s2_1) at (0, -0.9) {$y^{(2)}_1$};
          \node (s2_d) at (0.9, -0.9) {$\dots$};
          \node[rollout, suffix] (s2_N) at (1.9, -0.9) {$y^{(2)}_R$};
          \node[font=\scriptsize] (theta2) at (4.1, -0.9) {$\tilde p^{(2)}$ or $\tilde Q^{(2)}$};
          \draw[arrow] (s2_N) -- (theta2) node[midway, above=0.01cm, font=\tiny\sffamily] {count};

          \node at (0.9, -1.5) {$\vdots$};
          \node at (4.1, -1.5) {$\vdots$};

          \node[font=\sffamily\tiny, anchor=east] at (-0.4, -2.1) {Rollout $L$};
          \node[rollout, suffix] (sL_1) at (0, -2.1) {$y^{(L)}_1$};
          \node (sL_d) at (0.9, -2.1) {$\dots$};
          \node[rollout, suffix] (sL_N) at (1.9, -2.1) {$y^{(L)}_R$};
          \node[font=\scriptsize] (thetaL) at (4.1, -2.1) {$\tilde p^{(L)}$ or $\tilde Q^{(L)}$};
          \draw[arrow] (sL_N) -- (thetaL) node[midway, above=0.01cm, font=\tiny\sffamily] {count};

        \end{tikzpicture}
      }
      \caption{PMC schematic}
      \label{fig:pmc-sampling}
    \end{subfigure}%
    \quad
    \begin{subfigure}[t]{\widthof{\includegraphics[height=4cm]{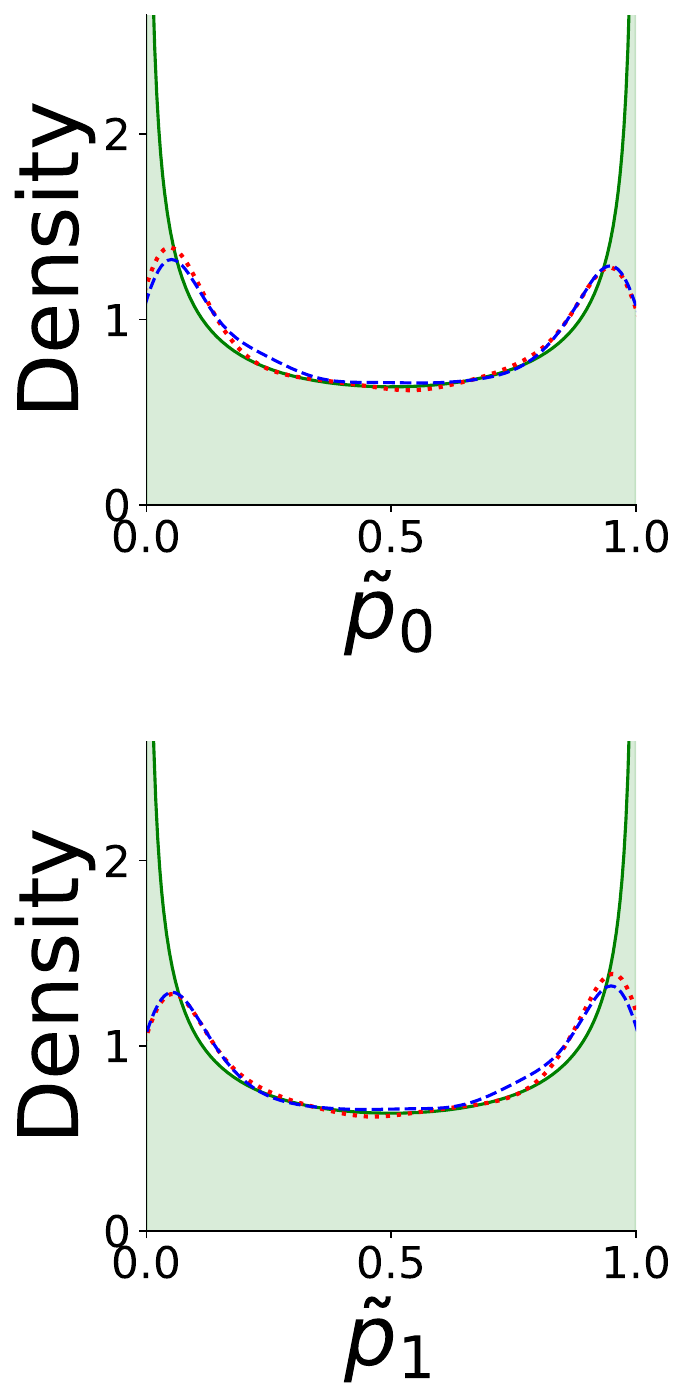}}}
      \centering
      \includegraphics[height=4cm]{figures/pmc_k0.pdf}
      \caption{Validation, $k=0$}
      \label{fig:pmc-k0}
    \end{subfigure}%
    \quad
    \begin{subfigure}[t]{\widthof{\includegraphics[height=4cm]{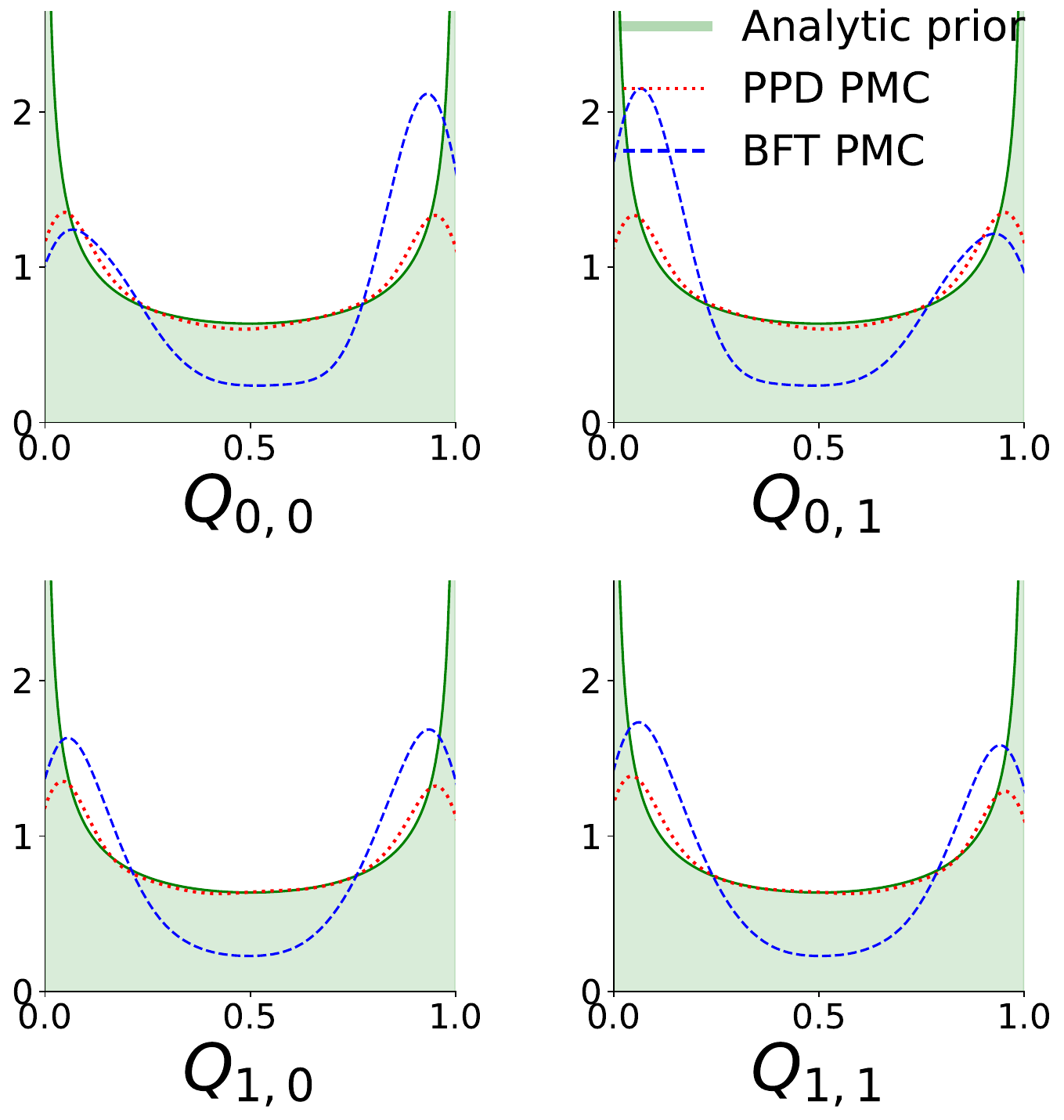}}}
      \centering
      \includegraphics[height=4cm]{figures/pmc_k1.pdf}
      \caption{Validation, $k=1$}
      \label{fig:pmc-k1}
    \end{subfigure}%
  }
  \caption{\textbf{Predictive Monte Carlo (PMC).} \textbf{(a)}~Schematic, adapted from \citet{ng2026tabmgpmartingaleposteriortabpfn}: each of $L$ rollouts autoregressively samples a length-$R$ sequence from $\pi_\theta$ with no conditioning; token frequencies give a prior sample $\tilde p^{(l)}$ ($k=0$) and transition counts give $\tilde Q^{(l)}$ ($k\geq 1$). \textbf{(b)--(c)}~Validation of the sampled prior density against the closed form, for $k=0$ and $k=1$. \emph{Analytic prior} (green, filled) is the closed-form prior $\Pi_0$; \emph{PPD PMC} (red, dotted) runs PMC on the exact posterior predictive (PPD) rule, isolating PMC's own sampling error; \emph{BFT PMC} (blue, dashed) runs PMC on the trained BFT, the sampler we actually use ($L{=}5000$ rollouts, $R{=}2000$).}
  \label{fig:pmc-validation}
\end{figure}

We maximize $J_{\mathrm{tilt}}$ by gradient ascent on this parameter. The gradient splits into two terms:
\begin{equation}
  \nabla J_{\mathrm{tilt}}(\kappa)
  = \Cov_{\tilde k\sim\Pi_{\mathrm{tilt}}(\cdot;\,\kappa)}\!\big(\bar\mu(\tilde k;\,\kappa),\,\nabla\log W(\tilde k;\,\kappa)\big)
  + \E_{\tilde k\sim\Pi_{\mathrm{tilt}}(\cdot;\,\kappa)}\!\big[\nabla\bar\mu(\tilde k;\,\kappa)\big],
  \label{eq:grad_Jtilt_general}
\end{equation}
where all gradients are with respect to this parameter. The first term is a covariance under $\Pi_{\mathrm{tilt}}$ between $\bar\mu$ and the score of $\log W$; the second vanishes when $\bar\mu$ does not depend on it. We estimate \eqref{eq:grad_Jtilt_general} by importance sampling against samples drawn once from the latent prior the BFT implies. These prior samples come from \emph{predictive Monte Carlo} (PMC), an instance of predictive Bayesian inference via martingale posteriors \citep{fongMartingalePosterior2023}, applied to BFTs by \citet{effiezalaswadi2026bft}: each sample is obtained by rolling out the BFT autoregressively without conditioning and reading the implied latent off the rollout's token statistics. Appendix~\ref{app:pmc} derives PMC for our setting and validates it against the analytic prior (Figure~\ref{fig:pmc-validation}). Because the samples are drawn once and reused across optimization steps, the optimization needs no further transformer calls and no backpropagation through $\pi_\theta$.

\textbf{Recovering a hard prompt.} The optimization above returns a kernel $\kappa$ from which prompts are sampled, but the elicitation problem asks for a hard prompt $z_{1:m}\in\cY^m$. We recover one by snapping the continuous optimum: for $k=0$ this means rounding the per-token sampling frequencies; for $k\geq 1$, constructing a hard prompt from the optimized transition table via an Eulerian path on its de~Bruijn graph. Both procedures are given in Appendix~\ref{app:hard_prompt}.

\subsection{The exchangeable case (\texorpdfstring{$k=0$}{k=0})}
\label{sec:ppt_k0}

Consider a BFT trained on i.i.d.\ sequences: the prior $\Pi_0$ is supported on $\cK_0$, the i.i.d.\ kernels of the form $\tilde k(y\mid h) = \tilde p(y)$ for some pmf $\tilde p \in \Delta(\cY)$. Identifying $\cK_0$ with $\Delta(\cY)$, the latent-posterior factorization \eqref{eq:icb-general} becomes a Bayes mixture over the latent pmf:
\begin{align}
  \pi_\theta(y \mid y_{1:n})  & = \int_{\Delta(\cY)} \tilde p(y)\, \Pi(d\tilde p \mid y_{1:n}),                &
  \Pi(d\tilde p \mid y_{1:n}) & \,\propto\, \Big[\textstyle\prod_{i=1}^n \tilde p(y_i)\Big]\,\Pi_0(d\tilde p).
  \label{eq:icb-k0}
\end{align}

\textbf{Instantiating the definitions in Table~\ref{tab:tilt-objects}.} The prompt kernel $\kappa\in\cK_0$ is determined by a pmf $\alpha\in\Delta(\cY)$ via $\kappa(y\mid h) = \alpha(y)$. The objects in Table~\ref{tab:tilt-objects} become functions of $\tilde p$ and $\alpha$. The \textbf{tilt factor} takes the form
\begin{equation}
  W(\tilde p;\, \alpha) = \E_{z_{1:m} \sim \alpha^{\otimes m}} \left[\prod_{j=1}^m \tilde p(z_j)\right]
  = \left[\sum_{v \in \cY} \alpha(v)\, \tilde p(v)\right]^m.
  \label{eq:W_k0}
\end{equation}
For $k=0$, $\mu(\tilde k;\,z_{1:m})$ does not depend on $z_{1:m}$ (the latent kernel ignores history); abbreviate
\begin{equation}
  \mu(\tilde p) \,:=\, \mu(\tilde k;\,z_{1:m}) \,=\, \E_{Y_{1:N}\sim\tilde p^{\otimes N}}[U(Y_{1:N})].
  \label{eq:mu_k0}
\end{equation}
The marginalized utility $\bar\mu(\tilde p;\,\alpha) = \mu(\tilde p)$ trivially.

\textbf{Score function gradient.}
The gradient of $J_{\mathrm{tilt}}$ specializes \eqref{eq:grad_Jtilt_general} (the second term vanishes since $\bar\mu = \mu(\tilde p)$ is $\alpha$-independent):
\begin{equation}
  \nabla_\alpha J_{\mathrm{tilt}}(\alpha)
  =
  \Cov_{\tilde p \sim \Pi_{\mathrm{tilt}}(\cdot;\, \alpha)}
  \big(\mu(\tilde p),\, \nabla_\alpha \log W(\tilde p;\, \alpha)\big).
  \label{eq:grad_Jtilt_k0}
\end{equation}
\textbf{IS estimator.}
We estimate the gradient \eqref{eq:grad_Jtilt_k0} by importance sampling from the BFT's latent prior. Since this prior does not depend on $\alpha$, we draw $L$ prior samples $\tilde p^{(\ell)}$ from it once, via PMC (Appendix~\ref{app:pmc}), and reuse them across all optimization steps.

Given the current $\alpha$, define the \textbf{importance weights} $w_\ell := W(\tilde p^{(\ell)};\,\alpha)$ and their self-normalized form $\tilde w_\ell := w_\ell / \sum_{r=1}^L w_r$, and let $\widehat\mu^{(\ell)}$ denote an estimator of $\mu(\tilde p^{(\ell)})$. We estimate $\nabla_{\alpha}J_{\mathrm{tilt}}(\alpha)$ by the self-normalized sample covariance:
\begin{equation}
  \widehat{\nabla_{\alpha}J_{\mathrm{tilt}}}(\alpha)
  :=\sum_{\ell=1}^L \tilde w_\ell\,
  \Big(\widehat{\mu}^{(\ell)}-\bar\mu_{w}\Big)\,
  \Big(s^{(\ell)}-\bar s_{w}\Big),
  \label{eq:IS_estimator}
\end{equation}
where
\[
  s^{(\ell)} := \nabla_\alpha \log W(\tilde p^{(\ell)};\,\alpha), \quad
  \bar\mu_{w} := \sum_\ell \tilde w_\ell \widehat{\mu}^{(\ell)}, \quad
  \bar s_{w} := \sum_\ell \tilde w_\ell s^{(\ell)}.
\]
The two variants of \eqref{eq:IS_estimator} differ only in how they estimate $\mu(\tilde p^{(\ell)})$. \textsc{PPT} uses a single rollout: it draws a continuation $Y^{(\ell)}_{1:N} \sim (\tilde p^{(\ell)})^{\otimes N}$ and sets $\widehat\mu^{(\ell)} := U(Y^{(\ell)}_{1:N})$, which is unbiased given $\tilde p^{(\ell)}$ but noisy. \textsc{PPT-RB} instead uses the closed-form $\mu(\tilde p^{(\ell)})$ from \eqref{eq:mu_k0}, available here (Appendix~\ref{app:closed_forms}). Replacing the rollout by its conditional expectation removes the rollout noise, so by Rao--Blackwell (Appendix~\ref{app:rb}) \textsc{PPT-RB} has lower variance.

\subsection{The \texorpdfstring{$k$}{k}-Markov exchangeable case (\texorpdfstring{$k \geq 1$}{k >= 1})}
\label{sec:ppt_kmarkov}

This subsection generalizes Section~\ref{sec:ppt_k0} to $k\geq 1$, with $k=0$ recovered as a special case. Consider a BFT trained on $k$-Markov exchangeable sequences: the prior $\Pi_0$ is supported on $\cK_k$, the $k$-Markov kernels of the form $\tilde k(y\mid h) = \tilde Q_{\mathrm{suf}_k(h),\, y}$ for some transition table $\tilde Q \in \cQ_k := \Delta(\cY)^{|\cY|^k}$. Identifying $\cK_k$ with $\cQ_k$, the latent-posterior factorization \eqref{eq:icb-general} becomes a Bayes mixture over the latent transition table:
\begin{align}
  \pi_\theta(y \mid y_{1:n})  & = \int_{\cQ_k} \tilde Q_{\mathrm{suf}_k(y_{1:n}), y}\, \Pi(d\tilde Q \mid y_{1:n}), &
  \Pi(d\tilde Q \mid y_{1:n}) & \propto \prod_{i=k+1}^n \tilde Q_{\mathrm{suf}_k(y_{1:i-1}), y_i} \Pi_0(d\tilde Q).
  \label{eq:icb-kmarkov}
\end{align}

\textbf{Instantiating the definitions in Table~\ref{tab:tilt-objects}.} The prompt is a Markov chain on $\cY^m$ with two parameters: a transition table $A\in\cQ_k$, where $A_{s,v}$ is the probability of emitting $v$ from state $s$, and a distribution $\rho\in\Delta(\cY^k)$ over the first $k$ tokens. We write $P_{A,\rho}$ for the resulting law on $\cY^m$: the first $k$ tokens are drawn from $\rho$, and each subsequent token $z_j$ from $A(\cdot \mid \mathrm{suf}_k(z_{1:j-1}))$. Equivalently, the prompt kernel is $\kappa(y\mid h) = A_{\mathrm{suf}_k(h),\,y}$. The objects in Table~\ref{tab:tilt-objects} become functions of $\tilde Q, A, \rho$. The \textbf{tilt factor} takes the form
\begin{equation}
  W(\tilde Q;\, A, \rho)
  = \E_{z_{1:m}\sim P_{A,\rho}}\!\!\left[\prod_{j=k+1}^m \tilde Q_{\mathrm{suf}_k(z_{1:j-1}),\, z_j}\right],
  \label{eq:W_kmarkov}
\end{equation}
where the product starts at $j=k+1$ because the first $k$ tokens have no preceding context to transition from; $\rho$ enters \eqref{eq:W_kmarkov} only through $P_{A,\rho}$, which determines the chain's starting state. Appendix~\ref{app:W_closed_form} gives a closed-form matrix-power expression for $W$, which we use throughout.

For $k$-Markov $\tilde k$, $\mu(\tilde k;\,z_{1:m})$ depends on $z_{1:m}$ only through its length-$k$ suffix; abbreviate
\begin{equation}
  \mu(\tilde Q;\, s) \,:=\, \mu(\tilde k;\,z_{1:m})\big|_{\mathrm{suf}_k(z_{1:m})=s} \,=\, \E_{Y_{1:N}\sim\tilde Q(\cdot\mid s)}[U(Y_{1:N})], \qquad s\in\cY^k.
  \label{eq:mu_kmarkov}
\end{equation}
Closed forms for $\mu(\tilde Q;\,s)$ in our utility/BFT pairs are in Appendix~\ref{app:closed_forms}. The marginalized utility is the average of $\mu(\tilde Q;\,s)$ over the distribution $\nu_s(A,\rho)$ that the suffix $\mathrm{suf}_k(z_{1:m})$ has under $P_{A,\rho}$:
\begin{equation}
  \bar\mu(\tilde Q;\, A, \rho) = \sum_{s \in \cY^k} \nu_s(A, \rho)\, \mu(\tilde Q;\, s),
  \label{eq:mu_weighted}
\end{equation}
Since $\nu_s(A,\rho)$ is itself closed form (Appendix~\ref{app:W_closed_form}), $\bar\mu$ is available in closed form whenever $\mu(\tilde Q;\,s)$ is, as for our utility/BFT pairs; \textsc{PPT-RB} uses this closed form.

\textbf{Score function gradient.}
The gradient with respect to $A$ has two terms, one through the tilt weights and one through $\nu_s(A,\rho)$:
\begin{equation}
  \nabla_A J_{\mathrm{tilt}}(A, \rho)
  =
  \Cov_{\tilde Q \sim \Pi_{\mathrm{tilt}}}\!\big(\bar\mu(\tilde Q;\, A, \rho),\, \nabla_A \log W(\tilde Q;\, A, \rho)\big)
  + \E_{\tilde Q \sim \Pi_{\mathrm{tilt}}}\!\big[\nabla_A \bar\mu(\tilde Q;\, A, \rho)\big].
  \label{eq:grad_Jtilt_kmarkov}
\end{equation}
The score $\nabla_A \log W$ is always available in closed form, by differentiating the matrix power \eqref{eq:W_matrix_power} (Appendix~\ref{app:W_closed_form}). The second term, $\nabla_A \bar\mu = \sum_s \mu(\tilde Q;\,s)\,\nabla_A \nu_s(A,\rho)$, is closed form whenever $\mu(\tilde Q;\,s)$ is (as for our utility/BFT pairs), which \textsc{PPT-RB} uses; otherwise \textsc{PPT} estimates the integrand by a single rollout. We optimize $\rho$ jointly with $A$ by simplex-projected gradient descent (Appendix~\ref{app:rho_update}).

The IS estimator extends \eqref{eq:IS_estimator}: the second term of \eqref{eq:grad_Jtilt_kmarkov} contributes an additional importance-weighted expectation, and the integrand $\bar\mu(\tilde Q^{(\ell)};\, A, \rho)$ can be evaluated either in closed form (\textsc{PPT-RB}) or by a single rollout (\textsc{PPT}). The full estimator is in Appendix~\ref{app:is_kmarkov}.

\section{Experiments}
\label{sec:experiments}
Our experiments use $\cY=\{0,1\}$, continuation horizon $N=4$, Markov orders $k\in\{0,1\}$, and prompt lengths $m\in\{6,50\}$. We meta-learn two BFTs. For the \emph{Beta--Bernoulli BFT} ($k=0$), $\Pi_0 = \mathrm{Beta}(1/2,1/2)$ on the latent pmf $\tilde p$ and tokens are conditionally i.i.d.\ $\mathrm{Bernoulli}(\tilde p)$; for the \emph{reinforced urn BFT} ($k=1$), $\Pi_0$ has independent rows $\tilde Q_{a,\cdot}\stackrel{\text{ind.}}{\sim}\mathrm{Dir}(1/2,1/2)$ on the latent transition matrix $\tilde Q\in\cQ_1$, and the sequence is a Markov chain with transition matrix $\tilde Q$. In both cases the posterior on the latent kernel and the posterior predictive on the next token are available in closed form (Appendices~\ref{app:beta_bernoulli_derivations} and~\ref{app:reinforced_urn_derivations}); architectures and training details are in Appendix~\ref{app:bft_training}.

\subsection{Utility functions}
\label{sec:utilities}

\textbf{Reverse cross-entropy against a target distribution.} Fix a target distribution $\xi^\star$ on $\cY^N$ and set $U_{\xi^\star}(y_{1:N}) := \log \xi^\star(y_{1:N})$. Then $J(z_{1:m})$ is the negative cross-entropy from $\pi_\theta(\cdot\mid z_{1:m})$ to $\xi^\star$, so maximizing $J$ rewards prompts whose continuation distribution concentrates on continuations that $\xi^\star$ assigns high probability. For the Beta--Bernoulli BFT we take $\xi^\star = \mathrm{Bernoulli}(\tau^\star)^{\otimes N}$ and sweep $\tau^\star \in \{0.1, 0.2, 0.3, 0.4, 0.6, 0.7, 0.8, 0.9\}$ (8 targets). For the reinforced urn BFT we take $\xi^\star$ to be the law of a 1-Markov chain with transition matrix $Q^\star$, swept over two families: a \emph{symmetric} family (\texttt{sym-}$r$) with $Q^\star_{0,0}=Q^\star_{1,1}=r$ for $r\in\{0.1, 0.2, 0.3, 0.4, 0.6, 0.7, 0.8, 0.9, 1.0\}$ (9 targets), and a \emph{random} family (\texttt{dir-}$s$) of $10$ matrices indexed by $s\in\{0,\ldots,9\}$, with rows drawn i.i.d.\ from $\mathrm{Dir}(1/2,1/2)$ under seed $s$. We omit $\tau^\star = 0.5$ and $r = 0.5$ because both correspond to uniform targets under which $J = -N\log 2$ for every prompt and method.

\textbf{Match target frequency.} Set $U_{q^\star}(y_{1:N}) := -\big(f(y_{1:N}) - q^\star\big)^2$, where $f(y_{1:N}):=\frac{1}{N}\sum_{t=1}^N y_t$ is the empirical $1$-frequency. We sweep $q^\star\in\{0,\,0.1,\ldots,1.0\}$ (11 targets); full specification in Appendix~\ref{app:utility_specs}.

\textbf{Dyck validity.} Identifying $0\equiv$``\texttt{(}'' and $1\equiv$``\texttt{)}'', set $U_{\mathrm{dyck}}(y_{1:N}) = 1$ if $y_{1:N}$ is a valid balanced-bracket sequence and $0$ otherwise; full specification in Appendix~\ref{app:utility_specs}.

\subsection{Prefix-tuning methods compared}
\label{sec:baselines}

We compare five methods. \textbf{\textsc{GCG}} \citep{zou2023universal} optimizes hard prompts $z_{1:m} \in \cY^m$ directly via gradient-guided coordinate substitution (Appendix~\ref{app:gcg_details}). \textbf{\textsc{PPT}} and \textbf{\textsc{PPT-RB}} (Section~\ref{sec:ppt}) optimize a continuous parameterization ($(A,\rho)$ or $\alpha$) of the prompt distribution and snap to a hard prompt at termination (Appendix~\ref{app:hard_prompt}), with prior samples drawn from the trained BFT via PMC (Appendix~\ref{app:pmc}); closed-form $\mu$ (Appendix~\ref{app:closed_forms}) is available for all three utilities under both BFTs, so \textsc{PPT-RB} is applicable throughout. As diagnostic baselines, \textbf{\textsc{PPT (analytic)}} and \textbf{\textsc{PPT-RB (analytic)}} run the same procedure with samples drawn directly from $\Pi_0$, isolating the contribution of PMC sampling error. Implementation details are in Appendix~\ref{app:experiments}.

\textbf{Computational cost.} \textsc{PPT} and \textsc{PPT-RB} make no BFT queries during optimization; their only BFT cost is the one-time PMC sampling of $L=5000$ rollouts of length $R=2000$, i.e.\ $LR$ forward passes, amortized across all utilities. The \textsc{(analytic)} variants query the BFT not at all. \textsc{GCG}, in contrast, queries the BFT at every step: $|\cY|^N=16$ forward passes to evaluate $J(z)$, a backward pass to rank candidate bitflips, and $|\cY|^N$ further passes for each of the $c\leq m$ bitflips it considers, for $O((1+c)\,|\cY|^N)$ queries per step.
\subsection{Evaluation and results}
\label{sec:eval}

For any hard prompt $z_{1:m}\in\cY^m$ the elicitation objective $J(z_{1:m}) = \mathbb{E}_{Y_{1:N}\sim\pi_\theta(\cdot\mid z_{1:m})}[U(Y_{1:N})]$ is computed exactly by enumerating the $|\cY|^N=16$ continuations through $\pi_\theta$; this serves as our gold-standard score. For $m=6$ we additionally enumerate all $|\cY|^m=64$ candidate prompts to obtain the global optimum $J_\mathrm{opt}=\max_{z\in\cY^m} J(z)$ and the rank of every prompt; for $m=50$ the prompt space has $\sim 10^{15}$ elements, so $J_\mathrm{opt}$ and rank are unavailable.

We compare the five methods of Section~\ref{sec:baselines} on the Beta--Bernoulli and reinforced urn BFTs and the three utilities of Section~\ref{sec:utilities}, at $m\in\{6,50\}$. Every (utility, prompt length, BFT) setting is run over $10$ random seeds and we report means with standard errors. Table~\ref{tab:rev-xent-urn} summarizes reverse cross-entropy on the reinforced urn BFT. Per-configuration tables for all utilities and BFTs are in Appendix~\ref{app:susan-summary}; ESS and PMC diagnostics for PPT and PPT-RB are in Appendix~\ref{app:pmc-ess}.

Overall, the comparison depends on the utility, BFT, and prompt length. For reverse cross-entropy, the \textsc{PPT} variants substantially outperform \textsc{GCG} on the reinforced urn at both prompt lengths; on Beta--Bernoulli, all methods reach the optimum at $m=6$, and at $m=50$ \textsc{GCG} and \textsc{PPT-RB} perform equally well and better than \textsc{PPT}. For frequency match, on Beta--Bernoulli all methods reach near-optimum at $m=6$ and \textsc{GCG} outperforms the \textsc{PPT} variants at $m=50$; on the reinforced urn, the \textsc{PPT} variants outperform \textsc{GCG} at $m=6$, and \textsc{PPT-RB} and \textsc{GCG} are comparable at $m=50$. For Dyck validity, on Beta--Bernoulli all four \textsc{PPT} variants reach the $m=6$ optimum and all methods produce $J\approx 0.12$ at $m=50$; on the reinforced urn at $m=6$, \textsc{PPT-RB} uniquely reaches the enumerated optimum on every seed, and at $m=50$ the picture inverts with \textsc{GCG} producing $J=0.61$ versus $J\approx 0.014$--$0.021$ for the four \textsc{PPT} variants. The remainder of this section focuses on reverse cross-entropy; per-cell breakdowns for frequency match and Dyck validity are in Appendix~\ref{app:susan-summary}.

\textbf{Reverse cross-entropy.} On Beta--Bernoulli at $m=6$, all methods achieve $J$ at the enumerated optimum for every $\tau^\star$. At $m=50$ on Beta--Bernoulli, \textsc{GCG} and \textsc{PPT-RB} match each other across all $\tau^\star$, while \textsc{PPT} is consistently worse (Table~\ref{tab:rev-xent-bb}). On the reinforced urn, Table~\ref{tab:rev-xent-urn} reports mean $J$ at both $m\in\{6, 50\}$: the \textsc{PPT} variants achieve $J$ substantially closer to the optimum than \textsc{GCG} on both the symmetric and random target families. At $m=6$, Table~\ref{tab:rev-xent-urn-rank} reports the rank of each snapped prompt out of $|\cY|^m=64$: the \textsc{PPT} variants reach rank $1$--$3$ on most cells while \textsc{GCG} averages rank $5$--$16$.

\begin{table}[h]
  \centering
  \caption{Reverse cross-entropy on the reinforced urn BFT. Each cell is the mean of $J$ over $n=10$ random seeds (max convention; higher better), with standard error in parentheses. Best method per $m$ in bold (within $10^{-3}$).}
  \label{tab:rev-xent-urn}
  \resizebox{\textwidth}{!}{%
    \footnotesize
    \begin{tabular}{l rrrrr rrrrr}
      \toprule
                       & \multicolumn{5}{c}{$m=6$} & \multicolumn{5}{c}{$m=50$}                                                                                                                                                                                                                          \\
      \cmidrule(lr){2-6} \cmidrule(lr){7-11}
                       & GCG                       & PPT-RB (A)                 & PPT-RB                   & PPT (A)                   & PPT                      & GCG                      & PPT-RB (A)               & PPT-RB                   & PPT (A)                  & PPT                      \\
      \midrule
      \texttt{sym-0.1} & $-2.51\,(0.29)$           & $\mathbf{-1.75}\,(0.01)$   & $-1.80\,(0.00)$          & $-1.76\,(0.01)$           & $-1.80\,(0.00)$          & $-1.48\,(0.10)$          & $\mathbf{-0.61}\,(0.00)$ & $-0.61\,(0.00)$          & $-0.62\,(0.02)$          & $-0.65\,(0.02)$          \\
      \texttt{sym-0.2} & $-2.21\,(0.18)$           & $\mathbf{-1.73}\,(0.01)$   & $-1.76\,(0.00)$          & $\mathbf{-1.73}\,(0.01)$  & $-1.76\,(0.00)$          & $-1.56\,(0.06)$          & $\mathbf{-1.01}\,(0.00)$ & $\mathbf{-1.01}\,(0.00)$ & $-1.02\,(0.01)$          & $-1.03\,(0.01)$          \\
      \texttt{sym-0.3} & $-2.23\,(0.11)$           & $\mathbf{-1.94}\,(0.01)$   & $-1.96\,(0.00)$          & $-1.95\,(0.01)$           & $-1.96\,(0.00)$          & $-1.83\,(0.04)$          & $\mathbf{-1.50}\,(0.00)$ & $\mathbf{-1.50}\,(0.00)$ & $\mathbf{-1.50}\,(0.00)$ & $\mathbf{-1.50}\,(0.00)$ \\
      \texttt{sym-0.4} & $-2.43\,(0.05)$           & $\mathbf{-2.29}\,(0.00)$   & $-2.30\,(0.00)$          & $-2.29\,(0.00)$           & $-2.30\,(0.00)$          & $-2.24\,(0.02)$          & $\mathbf{-2.08}\,(0.00)$ & $-2.22\,(0.14)$          & $\mathbf{-2.08}\,(0.00)$ & $\mathbf{-2.08}\,(0.00)$ \\
      \texttt{sym-0.6} & $-2.37\,(0.06)$           & $\mathbf{-2.18}\,(0.00)$   & $\mathbf{-2.18}\,(0.00)$ & $\mathbf{-2.18}\,(0.00)$  & $\mathbf{-2.18}\,(0.00)$ & $-2.28\,(0.03)$          & $\mathbf{-2.06}\,(0.00)$ & $\mathbf{-2.06}\,(0.00)$ & $\mathbf{-2.06}\,(0.00)$ & $\mathbf{-2.06}\,(0.00)$ \\
      \texttt{sym-0.7} & $-2.11\,(0.14)$           & $\mathbf{-1.71}\,(0.01)$   & $\mathbf{-1.71}\,(0.01)$ & $\mathbf{-1.71}\,(0.01)$  & $\mathbf{-1.71}\,(0.01)$ & $-1.93\,(0.06)$          & $\mathbf{-1.47}\,(0.00)$ & $\mathbf{-1.47}\,(0.00)$ & $\mathbf{-1.47}\,(0.00)$ & $\mathbf{-1.47}\,(0.00)$ \\
      \texttt{sym-0.8} & $-2.01\,(0.22)$           & $\mathbf{-1.36}\,(0.01)$   & $\mathbf{-1.36}\,(0.01)$ & $\mathbf{-1.36}\,(0.01)$  & $\mathbf{-1.36}\,(0.01)$ & $-1.71\,(0.09)$          & $\mathbf{-0.96}\,(0.00)$ & $\mathbf{-0.96}\,(0.00)$ & $\mathbf{-0.96}\,(0.00)$ & $\mathbf{-0.96}\,(0.00)$ \\
      \texttt{sym-0.9} & $-2.19\,(0.35)$           & $\mathbf{-1.17}\,(0.02)$   & $\mathbf{-1.17}\,(0.02)$ & $\mathbf{-1.17}\,(0.02)$  & $\mathbf{-1.17}\,(0.02)$ & $-1.72\,(0.15)$          & $\mathbf{-0.53}\,(0.00)$ & $\mathbf{-0.53}\,(0.00)$ & $\mathbf{-0.53}\,(0.00)$ & $\mathbf{-0.53}\,(0.00)$ \\
      \texttt{sym-1.0} & $-24.10\,(4.79)$          & $\mathbf{-10.48}\,(0.23)$  & $-10.62\,(0.22)$         & $\mathbf{-10.48}\,(0.23)$ & $-10.62\,(0.22)$         & $-17.75\,(2.01)$         & $\mathbf{-1.50}\,(0.01)$ & $\mathbf{-1.50}\,(0.01)$ & $\mathbf{-1.50}\,(0.01)$ & $\mathbf{-1.50}\,(0.01)$ \\
      \midrule
      \texttt{dir-0}   & $-3.47\,(0.42)$           & $\mathbf{-1.89}\,(0.00)$   & $-2.04\,(0.15)$          & $\mathbf{-1.89}\,(0.00)$  & $-2.04\,(0.15)$          & $-1.76\,(0.03)$          & $\mathbf{-1.70}\,(0.16)$ & $-1.94\,(0.20)$          & $-1.90\,(0.18)$          & $-2.01\,(0.19)$          \\
      \texttt{dir-1}   & $-2.49\,(0.18)$           & $\mathbf{-1.63}\,(0.00)$   & $\mathbf{-1.63}\,(0.00)$ & $\mathbf{-1.63}\,(0.00)$  & $\mathbf{-1.63}\,(0.00)$ & $-1.65\,(0.04)$          & $\mathbf{-1.30}\,(0.00)$ & $-1.48\,(0.18)$          & $-1.65\,(0.23)$          & $-1.66\,(0.24)$          \\
      \texttt{dir-2}   & $-2.12\,(0.18)$           & $\mathbf{-1.32}\,(0.00)$   & $\mathbf{-1.32}\,(0.00)$ & $\mathbf{-1.32}\,(0.00)$  & $\mathbf{-1.32}\,(0.00)$ & $-2.32\,(0.08)$          & $-1.10\,(0.15)$          & $\mathbf{-0.95}\,(0.00)$ & $-1.25\,(0.20)$          & $-1.26\,(0.20)$          \\
      \texttt{dir-3}   & $-2.00\,(0.20)$           & $\mathbf{-1.40}\,(0.04)$   & $\mathbf{-1.40}\,(0.04)$ & $\mathbf{-1.40}\,(0.04)$  & $\mathbf{-1.40}\,(0.04)$ & $-1.74\,(0.09)$          & $\mathbf{-1.02}\,(0.04)$ & $\mathbf{-1.02}\,(0.04)$ & $-1.02\,(0.04)$          & $\mathbf{-1.02}\,(0.04)$ \\
      \texttt{dir-4}   & $-2.93\,(0.18)$           & $-2.04\,(0.23)$            & $-2.10\,(0.37)$          & $-2.04\,(0.23)$           & $\mathbf{-2.04}\,(0.23)$ & $-2.72\,(0.09)$          & $-1.14\,(0.37)$          & $\mathbf{-0.64}\,(0.27)$ & $-1.61\,(0.37)$          & $-1.33\,(0.37)$          \\
      \texttt{dir-5}   & $-2.27\,(0.23)$           & $\mathbf{-1.75}\,(0.00)$   & $\mathbf{-1.75}\,(0.00)$ & $\mathbf{-1.75}\,(0.00)$  & $\mathbf{-1.75}\,(0.00)$ & $-1.36\,(0.05)$          & $\mathbf{-0.78}\,(0.00)$ & $\mathbf{-0.78}\,(0.00)$ & $-0.82\,(0.02)$          & $-0.80\,(0.01)$          \\
      \texttt{dir-6}   & $-2.24\,(0.20)$           & $-1.45\,(0.13)$            & $\mathbf{-1.37}\,(0.12)$ & $-1.45\,(0.13)$           & $\mathbf{-1.37}\,(0.12)$ & $-1.79\,(0.15)$          & $-1.19\,(0.19)$          & $\mathbf{-0.93}\,(0.18)$ & $-1.19\,(0.19)$          & $-1.19\,(0.19)$          \\
      \texttt{dir-7}   & $-2.27\,(0.37)$           & $\mathbf{-1.22}\,(0.01)$   & $\mathbf{-1.22}\,(0.01)$ & $-1.23\,(0.01)$           & $\mathbf{-1.22}\,(0.01)$ & $-1.92\,(0.15)$          & $-0.54\,(0.05)$          & $-0.54\,(0.05)$          & $-0.55\,(0.04)$          & $\mathbf{-0.52}\,(0.05)$ \\
      \texttt{dir-8}   & $-2.61\,(0.13)$           & $\mathbf{-2.45}\,(0.03)$   & $\mathbf{-2.45}\,(0.03)$ & $\mathbf{-2.45}\,(0.03)$  & $\mathbf{-2.45}\,(0.03)$ & $\mathbf{-2.13}\,(0.02)$ & $-2.21\,(0.08)$          & $-2.22\,(0.08)$          & $-2.20\,(0.07)$          & $-2.21\,(0.08)$          \\
      \texttt{dir-9}   & $-2.51\,(0.19)$           & $\mathbf{-1.82}\,(0.00)$   & $-1.93\,(0.11)$          & $\mathbf{-1.82}\,(0.00)$  & $-1.93\,(0.11)$          & $-1.74\,(0.03)$          & $\mathbf{-1.68}\,(0.12)$ & $\mathbf{-1.68}\,(0.12)$ & $-1.80\,(0.16)$          & $-1.90\,(0.17)$          \\
      \bottomrule
    \end{tabular}}
\end{table}

\section{Related Works}
\label{sec:related}

\textbf{The Bayesian view of ICL.} Our framework rests on the view that ICL admits an approximate Bayesian interpretation, in which the transformer's predictions arise from a posterior over latent predictive models updated by the context. \citet{xie2022explanationincontextlearningimplicit} introduced this view as an explanation of ICL, and \citet{panwar2024incontextlearningbayesianprism} developed it empirically across a range of synthetic settings. \citet{garg2022what} and \citet{akyurek2023learningalgorithmincontextlearning} validated the picture in in-context linear regression, showing that transformers trained on hierarchical data implement Bayesian predictors. The connection between meta-learning on hierarchical data and Bayesian posterior prediction was formalized by \citet{ortegaMetalearningSequentialStrategies2019a}, whose analysis underlies our notion of an idealized BFT. Building on the same idealization, \citet{fortini2026uncertainty} prove a predictive central limit theorem for BFTs and use it to decompose a trained BFT's predictive uncertainty into aleatoric and epistemic components, using only forward passes.

\textbf{Belief-state geometry of trained transformers.} The latent-posterior factorization \eqref{eq:icb-general} is a hypothesis about what the trained transformer represents internally. \citet{shai2024transformersbeliefstate} showed that transformers trained on hidden Markov processes linearly represent the Bayesian belief state in their residual stream, with geometry matching the mixed-state presentation of computational mechanics. \citet{piotrowski2025constrainedbeliefupdates} extended this to constrained belief updates that account for the simplices observed in trained models, and \citet{riechers2025nexttokenicl} showed that next-token pretraining on data with hidden structure implies ICL of that structure. These results indicate that the latent-posterior model is not merely a useful idealization: trained transformers can be observed to maintain and update a posterior over latent generators. Complementing these internal probes, \citet{effiezalaswadi2026bft} recover a trained BFT's implied prior and posterior over the latent task from generation alone, via the same predictive Monte Carlo procedure we use in Appendix~\ref{app:pmc}.

\textbf{Prefix tuning for BFTs.} Two recent works also study prompting of BFTs: \citet{genewein2026understanding} and \citet{wenliang2025promptinghardunderstandingprompts}, both of which prompt Beta--Bernoulli BFTs (the $k=0$ case in our paper) with different Beta hyperparameters than ours: \citet{genewein2026understanding} use $\alpha=\beta=1$ (uniform-bias coin pretraining), and \citet{wenliang2025promptinghardunderstandingprompts} sweep over $(\alpha,\beta)$. \citet{wenliang2025promptinghardunderstandingprompts} additionally consider two simpler Bernoulli BFTs whose latent prior is a point mass at a single coin bias or a two-point mixture. Neither work covers the $k\geq 1$ Markov case.

The prompt/prefix tuning problem as defined in \citet{genewein2026understanding} refers to aligning the BFT to a
\emph{target} sequence distribution $\xi_{\mathrm{Target}}(y_{1:N})$ by \emph{prepending} a learnable prefix
$z_{1:m}\in\mathcal S^m$, where the prefix alphabet $\mathcal S$ varies by method: HardPT uses $\mathcal S=\cY$, SimplexPT uses $\mathcal S=\Delta(\cY)\subset\mathbb R^{|\cY|}$, RealPT uses $\mathcal S=\mathbb R^{|\cY|}$, and SoftPT uses $\mathcal S=\mathbb R^{d_{\mathrm{embed}}}$ (``embedding-dimensionality''). Their optimization objective (Eq.~(7)), in our notation, is
\[
  \min_{z_{1:m}\in\mathcal S^m}\;
  \E_{y_{1:N}\sim \xi_{\mathrm{Target}}}\!\left[ -\log \pi_\theta(y_{1:N} \mid z_{1:m})\right].
\]
Note their objective is the ``forward'' (mass-seeking) cross-entropy under the target distribution, which they minimize; maximizing our $J$ is equivalent to minimizing the ``reverse'' (mode-seeking) cross-entropy.
\citet{wenliang2025promptinghardunderstandingprompts} share the BFT setting and the forward cross-entropy objective above, but their focus is empirical: they show that exhaustive search and intuitive prompts often fail to identify the theoretically optimal prompt, and that optimal prompts depend on the pretraining distribution in unintuitive ways.

\textbf{Hard-prompt elicitation and persona-modulation attacks.} Outside the Bayesian view, a parallel literature attacks the elicitation problem by direct hard-prompt search. \citet{zou2023universal} introduced Greedy Coordinate Gradient (GCG), which optimizes a hard prompt to elicit a target completion via gradient-guided coordinate substitution; we use GCG as our experimental baseline. A complementary line of work exploits the latent structure that PPT operates on directly. \citet{shah2023scalabletransferableblackboxjailbreaks} demonstrate that ``persona-modulation attacks,'' designed to steer a model into adopting a particular persona, are an effective and transferable jailbreak strategy, and \citet{deshpande2023toxicity} show that persona assignment systematically shifts a model's distribution of completions. From the latent-posterior viewpoint, both can be understood as engineering prompts that bias the latent posterior toward kernels with high adversarial utility.

\textbf{Personas as latent factors.} A growing body of work treats LLM behavior as conditioned on structured latent character traits, or \emph{personas}. The Persona Selection Model of \citet{marks2026persona} hypothesizes that the assistant character is one component of a richer latent posterior maintained by post-trained LLMs, and frames misalignment as undesirable persona selection. \citet{ghandeharioun2024whosaskinguserpersonas} similarly frame alignment in terms of personas, arguing that both the model's adopted persona and its inferred user persona shape responses. \citet{joshi2024personaswaymodeltruthfulness} treat personas as latent factors and use this framing to explain the emergence of a ``truth direction'' in the latent space of language models. Most recently, \citet{wang2026persona} exhibit persona features in fine-tuned models that mediate emergent misalignment under narrow fine-tuning. From our perspective, these strands of evidence point to $\cK$ as the locus of behavioral steering: elicitation reduces to placing the latent posterior on a kernel, persona, or other latent factor that produces the target behavior.

\section{Discussion}
\label{sec:discussion}

\textbf{Summary.} We formalized the elicitation problem for Bayes-filtered transformers and showed that, under the latent-posterior factorization \eqref{eq:icb-general}, a tilted surrogate $J_{\mathrm{tilt}}$ of the elicitation objective admits a gradient \eqref{eq:grad_Jtilt_general} estimable entirely in latent-prior space. Posterior Prefix Tuning (PPT) operationalizes this in 0- and 1-Markov exchangeable BFTs: the prompt is sampled from a kernel whose continuous parameter is a pmf on $\cY$ ($k=0$) or a transition matrix on $\cY$ ($k\geq 1$), and the gradient is estimated by importance sampling with samples drawn from the BFT's latent prior. These samples are obtained once via predictive Monte Carlo and reused across optimization steps and across utility functions, so each gradient step requires zero transformer calls and no backpropagation through $\pi_\theta$.

\textbf{Limitations.} The method requires \eqref{eq:icb-general} to hold for the trained transformer. Even in our stylized settings, where meta-learning data is drawn from a $k$-Markov exchangeable prior so the idealized log-loss optimum satisfies \eqref{eq:icb-general}, the trained BFT only approximates this representation, and real-world transformers approximate it at best loosely. The scope is limited in three further ways. First, PPT is restricted to the prefix-tuning setting; extending it to settings where the BFT additionally conditions on observed data is left for future work. Second, our experiments cover only $|\cY|=2$, $k\in\{0,1\}$, and continuation horizon $N=4$. Third, hard-prompt recovery is a separate procedure from the continuous optimization (floor/ceil candidate evaluation for $k=0$; Eulerian-path construction on the de Bruijn graph for $k=1$), and we do not establish that the recovered hard prompt is optimal among those consistent with the optimized $\kappa$.

\textbf{Outlook.} The PMC samples are utility-independent and can be reused across any number of utility functions; the cost of drawing them from the BFT via PMC is paid once. Extending PPT beyond $k$-Markov exchangeability to other latent classes with finite-dimensional parameterizations would broaden the settings in which the latent-posterior factorization holds. 

\bibliographystyle{plainnat}
\bibliography{references}

\appendix

\section{Markov exchangeability}
\label{app:markov-exch}

The representation \eqref{eq:icb-general} is deliberately broad: it posits that the
model's next-token distribution is a Bayes mixture over latent tasks
$\tilde k\in\cK$, where $\cK$ contains arbitrary history-dependent kernels and the posterior
$\Pi(\cdot\mid y_{1:n})$ ranges over an essentially infinite-dimensional object.

We restrict attention to structured subclasses in which the
latent task admits a \emph{finite-dimensional parameterization} and posterior updating depends
on \emph{low-dimensional sufficient statistics} of the observed history (typically token or $k$-gram
transition counts). Exchangeability ($k=0$) and $k$-Markov exchangeability provide exactly this structure:
by classical representation theorems, such processes are mixtures of i.i.d.\ models or mixtures of
$k$th-order Markov chains, so the latent object reduces to a random pmf (exchangeable case) or a random
transition table (Markov-exchangeable case). In the remainder of this section we recall these notions
and introduce the corresponding definitions of \emph{in-context exchangeable} and \emph{in-context
  $k$-Markov exchangeable} transformers, which form the stylized setting used throughout the paper.

Throughout this appendix, $\cY$ is a finite vocabulary and $(Y_n)_{n\ge1}$ is an infinite sequence with joint law $\mathbb P$; we write $\Delta(\cY)$ for the simplex of probability measures (pmfs) on $\cY$. The representation theorems we invoke require exchangeability---and, in Section~\ref{sec:1markov-exch}, Markov exchangeability---of this full infinite sequence, and finiteness of $\cY$ is what makes the relevant parameter spaces finite-dimensional: the simplex $\Delta(\cY)$ in the exchangeable case, and the set of row-stochastic transition matrices on $\cY$ (resp.\ on $\cY^k$) in the $1$- (resp.\ $k$-) Markov-exchangeable case.

\subsection{Exchangeability}\label{sec:exch}
The infinite sequence $(Y_n)_{n\ge 1}$ on $\cY$ with joint law $\mathbb P$ is \emph{(infinitely) exchangeable} if for every $n\ge1$ and every
permutation $\sigma$ of $\{1,\dots,n\}$,
\[
  (Y_1,\dots,Y_n)\stackrel{d}{=}(Y_{\sigma(1)},\dots,Y_{\sigma(n)}).
\]
By the de Finetti representation theorem, the joint law of such a process is a mixture of i.i.d.\ laws: there is a unique probability measure $\Pi_0$ on $\Delta(\cY)$ such that
\[
  \mathbb P(Y_{1:n}=y_{1:n}) = \int_{\Delta(\cY)} \Big(\prod_{i=1}^n \tilde p(y_i)\Big)\,\Pi_0(d\tilde p),
  \qquad y_{1:n}\in\cY^n,
\]
where the directing random measure $\tilde p$ is the almost-sure limit of the empirical distributions $\tfrac1n\sum_{i=1}^n\delta_{Y_i}$. The one-step predictive distribution is
\[
  \mathbb P(Y_{n+1}\in A\mid Y_{1:n}=y_{1:n})
  =
  \int_{\Delta(\cY)} \tilde p(A)\,\Pi(d\tilde p\mid y_{1:n}),
  \qquad A\subseteq\cY,
\]
where $\Pi(\cdot\mid y_{1:n})$ is the posterior induced by the likelihood
$\prod_{i=1}^n \tilde p(y_i)$.
This is a special case of \eqref{eq:icb-general} obtained by restricting $\Pi_0$
to the $0$-Markov subclass $\cK_0$, i.e.\ kernels of the form $\tilde k(\cdot\mid h)\equiv \tilde p(\cdot)$.

The one-step conditionals of $\pi_\theta$ define consistent finite-dimensional joints through the chain rule, $\mathbb P_\theta(y_{1:n})=\prod_{i=1}^n\pi_\theta(y_i\mid y_{1:i-1})$, and these extend to a unique law $\mathbb P_\theta$ on the infinite-sequence space $\cY^{\mathbb N}$ (Ionescu--Tulcea, which for finite $\cY$ applies without further regularity conditions). We shall say the transformer $\pi_\theta$ is \textbf{in-context exchangeable}, or equivalently \textbf{in-context 0-Markov exchangeable}, if $\mathbb P_\theta$ is exchangeable; only then does the de Finetti representation above apply to it.

\subsection{1-Markov exchangeability}\label{sec:1markov-exch}
For a sequence $y_{1:n}\in\cY^n$, let
\[
  T^{(n)}_{a,b}(y_{1:n})
  :=
  \sum_{t=1}^{n-1} \mathbf 1\{y_t=a,\ y_{t+1}=b\},
  \qquad a,b\in\cY,
\]
count its $a\!\to\!b$ transitions. The infinite sequence $(Y_n)_{n\ge1}$ on $\cY$ is \emph{Markov exchangeable} if its law is invariant under reorderings of the steps that preserve the initial symbol and all transition counts: for every $n$ and all $y_{1:n},y'_{1:n}\in\cY^n$,
\[
  y_1=y'_1
  \ \text{and}\
  T^{(n)}_{a,b}(y_{1:n})=T^{(n)}_{a,b}(y'_{1:n})\ \forall a,b
  \ \Longrightarrow\
  \mathbb P(Y_{1:n}=y_{1:n})=\mathbb P(Y_{1:n}=y'_{1:n}).
\]

Let
\[
  \cQ
  :=
  \Big\{Q=(Q_{a,b})_{a,b\in\cY}:\ Q_{a,b}\in[0,1],\ \sum_{b\in\cY}Q_{a,b}=1\ \ \forall a\in\cY\Big\}
\]
denote the set of row-stochastic transition matrices on $\cY$, write $Q_a(A):=\sum_{b\in A}Q_{a,b}$ for $a\in\cY,\,A\subseteq\cY$, and call $(Y_n)$ \emph{recurrent} if its initial state $Y_1$ is $\mathbb P$-a.s.\ visited infinitely often. The analogue of de Finetti's theorem here is the Diaconis--Freedman representation of a recurrent Markov-exchangeable process as a mixture of Markov chains \citep[Theorem~4.6]{fortiniExchangeabilityPredictionPredictive2025}: there is a unique prior $\Pi^{\cQ}_0$ on $\cQ$ such that, for every $y_{1:n}$,
\begin{equation}
  \mathbb P(Y_{2:n}=y_{2:n}\mid Y_1=y_1)
  =
  \int_{\cQ} \Big(\prod_{t=1}^{n-1} Q_{y_t,y_{t+1}}\Big)\,\Pi^{\cQ}_0(dQ).
  \label{eq:markov-mixture-joint}
\end{equation}
The directing random matrix $\tilde Q\sim\Pi^{\cQ}_0$ is recovered $\mathbb P$-a.s.\ as the entrywise limit of the normalized transition counts $\hat T^{(n)}_{a,b}:=T^{(n)}_{a,b}\big/\sum_{c\in\cY}T^{(n)}_{a,c}$ (set to $0$ when the denominator vanishes). Updating $\Pi^{\cQ}_0$ through the Markov likelihood $L_n(Q;y_{1:n}):=\prod_{t=1}^{n-1}Q_{y_t,y_{t+1}}$ gives the posterior $\Pi^{\cQ}(dQ\mid y_{1:n})\propto L_n(Q;y_{1:n})\,\Pi^{\cQ}_0(dQ)$, and the one-step predictive distribution is the posterior mixture
\begin{equation}
  \mathbb P(Y_{n+1}\in A\mid Y_{1:n}=y_{1:n})
  =
  \int_{\cQ} Q_{y_n}(A)\,\Pi^{\cQ}(dQ\mid y_{1:n}),
  \qquad A\subseteq\cY.
  \label{eq:markov-mixture-predictive}
\end{equation}

This is a special case of \eqref{eq:icb-general} under the $1$-Markov restriction $\tilde k(y\mid h)=\tilde Q_{\mathrm{suf}_1(h),\,y}$, i.e.\ when the latent kernel depends on the history only through its last symbol $\mathrm{suf}_1(h)$. Thus we shall say the transformer $\pi_\theta$ is \textbf{in-context 1-Markov exchangeable} if the joint law it induces through its in-context predictions $\mathbb P_\theta$ is $1$-Markov exchangeable.

\subsection{\texorpdfstring{$k$}{k}-Markov exchangeability}
Fix $k\ge 1$ and define the $k$-gram (block) process $S_t:=(Y_{t-k+1},\dots,Y_t)\in\cY^k$ for $t\ge k$.
We call $(Y_t)$ \emph{$k$-Markov exchangeable} if the block process $(S_t)_{t\ge k}$ is Markov exchangeable
in the sense of Section~\ref{sec:1markov-exch} (with state space $\cY^k$). The block chain is constrained:
$S_t=(y_{t-k+1},\dots,y_t)$ can transition only to states $(y_{t-k+2},\dots,y_t,y_{t+1})$ that share its length-$(k{-}1)$
suffix, so $(S_t)$ moves on the subgraph of admissible $k$-gram transitions. When $(S_t)$ is recurrent on this
subgraph (its initial $k$-gram $S_k$ is $\mathbb P$-a.s.\ revisited infinitely often), the Diaconis--Freedman
representation of Section~\ref{sec:1markov-exch}, applied to $(S_t)$, yields a mixture of Markov chains on
$\cY^k$, equivalently a mixture of $k$th-order Markov chains on $\cY$.

Equivalently, \eqref{eq:icb-general} holds with $\Pi_0$ supported on $\cK_k$,
i.e.\ $\tilde k(\cdot\mid h)$ depends on $h$ only through $\mathrm{suf}_k(h)$.
Thus we shall say the transformer $\pi_\theta$ is \textbf{in-context $k$-Markov exchangeable} if the joint law it induces through
its in-context predictions $\mathbb P_\theta$ is $k$-Markov exchangeable.

\section{Methodology details}
\label{app:methodology}

\subsection{Predictive Monte Carlo}
\label{app:pmc}

PPT and PPT-RB require samples from the latent prior that the trained BFT $\pi_\theta$ carries over the predictive object---$\tilde p$ in the $0$-Markov exchangeable case, $\tilde Q$ in the $k$-Markov exchangeable case. We obtain them by \emph{Predictive Monte Carlo} (PMC) \citep{fongMartingalePosterior2023, fortiniExchangeabilityPredictionPredictive2025, effiezalaswadi2026bft}, using only forward passes through $\pi_\theta$.

It is the BFT's \emph{own} implied prior that PMC samples, not the data-generating prior $\Pi_0$. The BFT is meta-trained on sequences drawn from $\Pi_0$ and the likelihood, but once trained it induces its own predictive law and, with it, an implied latent prior (and posterior) over the predictive object. An idealized BFT that exactly satisfies \eqref{eq:icb-general} makes this implied prior coincide with $\Pi_0$; a trained BFT only approximates it, so the PMC samples inherit the approximation error of $\pi_\theta$.

PMC is not a generic procedure for arbitrary sequence models: its validity rests on conditions on the model's predictive law under which the rollout statistics converge to a well-defined latent. \citet{effiezalaswadi2026bft} review such sufficient conditions and show empirically that, for BFTs of the kind used here, PMC recovers both the implied prior and posterior; in this paper we use only prior samples.

\paragraph{Procedure (PMC for a BFT meta-trained on $k$-Markov exchangeable processes).}
For each rollout $l=1,\dots,L$:
\begin{enumerate}
  \item Autoregressively sample a length-$R$ sequence from $\pi_\theta$ with no conditioning:
        \[
          y^{(l)}_{t} \sim \pi_\theta(\cdot \mid y^{(l)}_{1:t-1}), \qquad t=1,\dots,R.
        \]
  \item Count:
        \begin{itemize}
          \item \emph{$0$-Markov exchangeable}: the token frequencies $\tilde p^{(l)}(y) = \frac{\#\{t : y^{(l)}_t = y,\; 1\le t\le R\}}{R}$ for each $y\in\cY$.
          \item \emph{$k$-Markov exchangeable}: the empirical transition frequencies from $y^{(l)}_{1:R}$, normalized to obtain $\tilde Q^{(l)}$.
        \end{itemize}
\end{enumerate}
The collection $\{\tilde p^{(l)}\}_{l=1}^L$ (resp.\ $\{\tilde Q^{(l)}\}_{l=1}^L$) is a set of approximate samples from the BFT's implied latent prior. For the idealized BFT, whose predictive law is exactly the exchangeable (resp.\ $k$-Markov exchangeable) mixture \eqref{eq:icb-general}, each rollout's empirical frequencies converge almost surely as $R\to\infty$ to its directing latent---a draw from $\Pi_0$ \citep{fortiniExchangeabilityPredictionPredictive2025}. Two distinct errors separate this idealization from practice: the finite-rollout counting error, which $R$ controls and which vanishes as $R\to\infty$; and the gap between the trained $\pi_\theta$ and the idealized BFT, a fixed property of the model that no choice of $R$ reduces.

Figure~\ref{fig:pmc-sampling} illustrates the procedure. The key feature is that PMC requires only \emph{forward passes} through $\pi_\theta$ (no gradients, no internal access), making it applicable even with API-only access to the model.


\subsection{Closed-form computation of \texorpdfstring{$W$}{W} and its score}
\label{app:W_closed_form}

\paragraph{Exchangeable case.} The tilt factor \eqref{eq:W_k0} is $W(\tilde p;\,\alpha) = [\sum_v \alpha(v)\tilde p(v)]^m$. For $\cY=\{0,1\}$ the score with respect to $\alpha$ has the closed form
\[
  \nabla_\alpha \log W(\tilde p;\, \alpha) = \frac{m\,[\tilde p - (1-\tilde p)]}{\alpha\tilde p + (1-\alpha)(1-\tilde p)}.
\]

\paragraph{Markov-exchangeable case.} The tilt factor \eqref{eq:W_kmarkov} is an expectation over the prompt Markov chain on $\cY^m$. Changing variables from token sequences to state sequences gives a closed-form matrix-power expression:
\begin{equation}
  W(\tilde Q;\, A, \rho)
  = \rho^\top \bigl(M^{(\tilde Q)}\bigr)^{m-k}\, \mathbf{1},
  \label{eq:W_matrix_power}
\end{equation}
where $M^{(\tilde Q)} \in \mathbb{R}^{|\cY|^k \times |\cY|^k}$ has entries
\begin{equation}
  M^{(\tilde Q)}_{s,s'} = \sum_{v \in \cY} A_{s,v}\, \tilde Q_{s,v}\, \mathbf{1}[\mathrm{suf}_k(s,v) = s'].
  \label{eq:M_def}
\end{equation}
That is, $M^{(\tilde Q)} = A \odot \tilde Q$ in the Hadamard sense, after accounting for the suffix map. The cost is $O((m{-}k) \cdot |\cY|^{k+1} \cdot L)$ across all $L$ prior samples. When $k=0$, $M$ is $1 \times 1$ with $M = \sum_v \alpha(v) \tilde p(v)$, $\rho$ is trivial, and \eqref{eq:W_matrix_power} recovers \eqref{eq:W_k0}.

\paragraph{Gradient $\nabla_A \log W$.} Differentiating \eqref{eq:W_matrix_power} through the matrix power gives
\[
  \nabla_A \log W(\tilde Q;\, A, \rho) = \frac{\nabla_A \big[\rho^\top (M^{(\tilde Q)})^{m-k}\, \mathbf{1}\big]}{\rho^\top (M^{(\tilde Q)})^{m-k}\, \mathbf{1}},
\]
where the dependence on $A$ enters only through $M^{(\tilde Q)}$, with entrywise derivative
\[
  \frac{\partial M^{(\tilde Q)}_{s,s'}}{\partial A_{s,v}} = \tilde Q_{s,v}\,\mathbf{1}[\mathrm{suf}_k(s,v)=s'].
\]
The matrix-chain derivative is then obtained by differentiating through the matrix power.

\paragraph{Suffix distribution $\nu_s(A,\rho)$.} The distribution that $\mathrm{suf}_k(z_{1:m})$ has under $P_{A,\rho}$ is given by
\[
  \nu_s(A,\rho) = [\rho^\top T_A^{m-k}]_s,
\]
where $T_A \in \mathbb{R}^{|\cY|^k \times |\cY|^k}$ is the state transition matrix induced by $A$, with entries
\[
  [T_A]_{s,s'} = \sum_{v\in\cY} A_{s,v}\, \mathbf{1}[\mathrm{suf}_k(s,v) = s'].
\]
The gradient $\nabla_A \bar\mu$ in \eqref{eq:grad_Jtilt_kmarkov} is computed by differentiating $\rho^\top T_A^{m-k}$ through $T_A$.


\subsection{Rao--Blackwellization of the gradient estimator}
\label{app:rb}

For each prior sample $\tilde k^{(\ell)}$ from the BFT's latent prior, the IS estimator requires an estimate of the integrand $\bar\mu(\tilde k^{(\ell)};\,\kappa)$ (and, for $k\geq 1$, of $\nabla\bar\mu$). \textsc{PPT-RB} uses the closed-form $\bar\mu$; \textsc{PPT} uses a single Monte Carlo rollout,
\[
  z_{1:m}\sim\kappa, \qquad Y_{1:N}\sim\tilde k^{(\ell)}(\cdot\mid \mathrm{suf}_k(z_{1:m})), \qquad \widehat{\bar\mu}^{(\ell)} := U(Y_{1:N}).
\]

\paragraph{Variance reduction.} The rollout estimate is unbiased given $\tilde k^{(\ell)}$:
\[
  \E\!\left[\widehat{\bar\mu}^{(\ell)} \,\big|\, \tilde k^{(\ell)}\right] = \bar\mu(\tilde k^{(\ell)};\,\kappa).
\]
The law of total variance gives
\begin{equation}
  \Var\!\left(\widehat{\bar\mu}^{(\ell)}\right) \,=\, \Var\!\left(\bar\mu(\tilde k^{(\ell)};\,\kappa)\right) + \E\!\left[\Var\!\left(\widehat{\bar\mu}^{(\ell)} \,\big|\, \tilde k^{(\ell)}\right)\right],
  \label{eq:rb_lotv}
\end{equation}
so $\Var(\bar\mu(\tilde k^{(\ell)};\,\kappa)) \leq \Var(\widehat{\bar\mu}^{(\ell)})$, with the gap equal to the within-kernel sampling variance $\E[\Var(\widehat{\bar\mu}^{(\ell)}\mid\tilde k^{(\ell)})]$. Substituting the closed-form $\bar\mu$ for $\widehat{\bar\mu}^{(\ell)}$ in the IS estimator (\eqref{eq:IS_estimator} for $k=0$, \eqref{eq:IS_estimator_kmarkov} for $k\geq 1$) reduces the variance of each summand, hence the variance of the gradient estimator. The reduction is largest when individual rollouts from a fixed $\tilde k^{(\ell)}$ are noisy.

\paragraph{Specialization to $k=0$.} The kernel ignores history, so $\bar\mu(\tilde p;\,\alpha) = \mu(\tilde p)$ trivially and the rollout estimate is $\widehat\mu^{(\ell)} = U(Y^{(\ell)}_{1:N})$ with $Y^{(\ell)}_{1:N}\sim(\tilde p^{(\ell)})^{\otimes N}$ (no prompt sampling needed); \textsc{PPT-RB} uses the closed-form $\mu(\tilde p^{(\ell)})$.

\paragraph{When the closed form exists.} Closed-form expressions for $\mu(\tilde p)$ and $\bar\mu(\tilde Q;\, A, \rho)$ depend on the utility/BFT combination; see Appendices~\ref{app:beta_bernoulli_derivations} and \ref{app:reinforced_urn_derivations} for the cases used in our experiments. For utilities where no closed form is available (e.g., a black-box behavioral classifier), only \textsc{PPT} applies.


\subsection{IS gradient estimator for \texorpdfstring{$k\geq 1$}{k>=1}}
\label{app:is_kmarkov}

For $k\geq 1$ the score function gradient \eqref{eq:grad_Jtilt_kmarkov} has two terms; we estimate it by importance sampling using $L$ prior samples $\tilde Q^{(\ell)}$ drawn once via PMC from the BFT's latent prior (Appendix~\ref{app:pmc}) and reused across optimization steps. Given the current $(A,\rho)$, define the importance weights
\[
  w_\ell := W(\tilde Q^{(\ell)};\, A, \rho),
  \qquad
  \tilde w_\ell := \frac{w_\ell}{\sum_{r=1}^L w_r}.
\]
The estimator is a self-normalized sample covariance plus an importance-weighted expectation:
\begin{equation}
  \widehat{\nabla_A J_{\mathrm{tilt}}}(A, \rho)
  := \sum_{\ell=1}^L \tilde w_\ell\, \big(\widehat{\bar\mu}^{(\ell)} - \bar{\bar\mu}_w\big)\big(s^{(\ell)} - \bar s_w\big)
  \;+\; \sum_{\ell=1}^L \tilde w_\ell\, \nabla_A \bar\mu(\tilde Q^{(\ell)};\, A, \rho),
  \label{eq:IS_estimator_kmarkov}
\end{equation}
where
\[
  s^{(\ell)} := \nabla_A \log W(\tilde Q^{(\ell)};\, A, \rho), \qquad
  \bar{\bar\mu}_w := \sum_\ell \tilde w_\ell \widehat{\bar\mu}^{(\ell)}, \qquad
  \bar s_w := \sum_\ell \tilde w_\ell s^{(\ell)}.
\]
The integrand $\bar\mu(\tilde Q^{(\ell)};\, A, \rho)$ selects the variant. \textsc{PPT-RB} uses the closed form \eqref{eq:mu_weighted}, while \textsc{PPT} uses a single rollout,
\[
  z^{(\ell)}\sim P_{A,\rho}, \qquad
  Y^{(\ell)}_{1:N} \sim \tilde Q^{(\ell)}(\cdot \mid \mathrm{suf}_k(z^{(\ell)})), \qquad
  \widehat{\bar\mu}^{(\ell)} = U(Y^{(\ell)}_{1:N}),
\]
which is unbiased given $\tilde Q^{(\ell)}$,
\[
  \E\!\left[\widehat{\bar\mu}^{(\ell)}\mid\tilde Q^{(\ell)}\right] = \bar\mu(\tilde Q^{(\ell)};\, A, \rho).
\]
By Rao--Blackwell (Appendix~\ref{app:rb}), \textsc{PPT-RB} has strictly lower variance for each $\ell$.


\subsection{Updating the initial-state distribution \texorpdfstring{$\rho$}{rho}}
\label{app:rho_update}

The prompt law $P_{A,\rho}$ of Section~\ref{sec:ppt_kmarkov} has two parameters: the transition table $A\in\cQ_k$ and the initial-state distribution $\rho\in\Delta(\cY^k)$ over the first $k$ tokens. Section~\ref{sec:ppt_kmarkov} gives the update for $A$; here we describe the update for $\rho$, which we optimize by \emph{projected gradient ascent} on the simplex $\Delta(\cY^k)$, interleaved with the $A$ step.

\paragraph{$J_{\mathrm{tilt}}$ is linear in $\rho$.}
Fix $A$ and the $L$ prior samples $\tilde Q^{(\ell)}$, and hold the self-normalized importance weights $\tilde w_\ell \propto W(\tilde Q^{(\ell)};\,A,\rho)$ at their current values while differentiating in $\rho$. (The weights do depend on $\rho$ through $W$; freezing them gives a partial gradient. Unlike the $A$ gradient~\eqref{eq:grad_Jtilt_kmarkov}, which retains the covariance term through $\nabla_A\log W$, the $\rho$ step keeps only the term below. This is the gradient the implementation uses, and it suffices because $\rho$ is updated jointly with $A$ at every step.) With the weights fixed, the surrogate objective is the importance-weighted marginalized utility
\[
  J_{\mathrm{tilt}}(A,\rho) \;=\; \sum_{\ell=1}^L \tilde w_\ell\;\bar\mu(\tilde Q^{(\ell)};\,A,\rho).
\]
Substitute the marginalized utility~\eqref{eq:mu_weighted}, $\bar\mu(\tilde Q^{(\ell)};A,\rho)=\sum_{s\in\cY^k}\nu_s(A,\rho)\,\mu(\tilde Q^{(\ell)};s)$, together with the closed form $\nu_s(A,\rho)=[\rho^\top T_A^{m-k}]_s$ from Appendix~\ref{app:W_closed_form}:
\begin{equation}
  J_{\mathrm{tilt}}(A,\rho)
  \;=\; \sum_{\ell} \tilde w_\ell \sum_{s\in\cY^k} \big[\rho^\top T_A^{m-k}\big]_s\,\mu(\tilde Q^{(\ell)};s)
  \;=\; \rho^\top T_A^{m-k}\, \boldsymbol\mu_w,
  \qquad
  [\boldsymbol\mu_w]_s := \sum_{\ell} \tilde w_\ell\,\mu(\tilde Q^{(\ell)};s).
  \label{eq:Jtilt_rho_linear}
\end{equation}
Here $T_A^{m-k}$ propagates the starting state forward the $m-k$ transitions of the prompt, and $\boldsymbol\mu_w\in\mathbb R^{|\cY|^k}$ collects the importance-weighted utility at each possible ending state $s$. For fixed $A$ and frozen weights, $J_{\mathrm{tilt}}$ is therefore a \emph{linear} function of $\rho$, and its gradient is the constant vector
\begin{equation}
  \nabla_\rho J_{\mathrm{tilt}}(A,\rho) \;=\; T_A^{m-k}\,\boldsymbol\mu_w .
  \label{eq:grad_rho}
\end{equation}
There is no transpose: writing $J_{\mathrm{tilt}}=\rho^\top\big(T_A^{m-k}\boldsymbol\mu_w\big)$ makes clear that $\partial J_{\mathrm{tilt}}/\partial\rho_s=\big[T_A^{m-k}\boldsymbol\mu_w\big]_s$. Component $s$ of \eqref{eq:grad_rho} is the utility a prompt would accrue if it started in state $s$, averaged over the tilted posterior; gradient ascent thus shifts mass toward starting states with high expected utility.

\paragraph{Projected gradient ascent on the simplex.}
A bare ascent step $\rho + \eta\,\nabla_\rho J_{\mathrm{tilt}}$ would leave the simplex $\Delta(\cY^k)$: the updated vector need neither sum to one nor stay nonnegative. We correct the step in two stages, with learning rate $\eta$.

\emph{(i) Project the gradient onto the simplex tangent.} The simplex lies in the affine hyperplane $\{x:\sum_s x_s = 1\}$, whose tangent space is $\{v:\sum_s v_s = 0\}$: moving along any such $v$ leaves the coordinate sum unchanged. We project the gradient onto this tangent by subtracting its mean,
\[
  \widetilde g \;:=\; \nabla_\rho J_{\mathrm{tilt}} - \overline{\nabla_\rho J_{\mathrm{tilt}}}\,\mathbf 1,
  \qquad
  \overline{\nabla_\rho J_{\mathrm{tilt}}} := \frac{1}{|\cY|^k}\sum_{s}\big[\nabla_\rho J_{\mathrm{tilt}}\big]_s ,
\]
so that $\sum_s \widetilde g_s = 0$ and the step $\rho + \eta\,\widetilde g$ preserves $\sum_s \rho_s = 1$ to first order.

\emph{(ii) Project back into the simplex.} The tangent step can still drive components below zero, so afterward we map the result back into the simplex \emph{interior} by flooring at $\epsilon$ and renormalizing,
\[
  \mathrm{proj}_\Delta(x)_s \;:=\; \frac{\max(x_s,\,\epsilon)}{\sum_{s'}\max(x_{s'},\,\epsilon)},
  \qquad \epsilon = 10^{-6}.
\]
The floor keeps every $\rho_s \ge \epsilon$ after renormalization, preventing $\rho$ from collapsing onto a vertex of the simplex; a vertex would pin the prompt to a single deterministic starting state and stall the joint $(A,\rho)$ search. This clamp-and-renormalize map is a cheap stand-in for the exact Euclidean projection onto $\Delta(\cY^k)$, which we did not find necessary.

Combining the two stages, each update is
\begin{equation}
  \rho \;\leftarrow\; \mathrm{proj}_\Delta\!\Big(\rho \,+\, \eta\,\big(\nabla_\rho J_{\mathrm{tilt}} - \overline{\nabla_\rho J_{\mathrm{tilt}}}\big)\Big).
  \label{eq:rho_update}
\end{equation}
We interleave \eqref{eq:rho_update} with the $A$ update of Section~\ref{sec:ppt_kmarkov} at the same learning rate, so that $A$ and $\rho$ ascend $J_{\mathrm{tilt}}$ jointly. (The optimizer minimizes the stored objective $-J_{\mathrm{tilt}}$; \eqref{eq:rho_update} states the equivalent ascent on $J_{\mathrm{tilt}}$ in the paper's convention.)

\subsection{Hard prompt recovery}
\label{app:hard_prompt}

After PPT optimization, we recover a hard prompt $z_{1:m} \in \cY^m$ from the optimized prompt-distribution parameter as follows.

\paragraph{Exchangeable case ($k=0$).}
For binary $\cY = \{0,1\}$, evaluate $J_{\mathrm{tilt}}$ at both $h = \lfloor m\alpha \rfloor$ and $h = \lceil m\alpha \rceil$ ones, select the count with higher $J_{\mathrm{tilt}}$, and construct the prompt with that many 1s.

\paragraph{Markov-exchangeable case ($k \geq 1$).}
Convert the optimized $(A, \rho)$ to a hard prompt $z_{1:m}$ via an Eulerian path in the de~Bruijn graph. For each of the $|\cY|^k$ starting states $s_0$:
\begin{enumerate}
  \item Compute expected visit counts $\nu(u) = \sum_{j=0}^{m-k-1} [e_{s_0}^\top T_A^j]_u$ for each $k$-gram state $u$.
  \item Assign edge multiplicities via Hamilton's method (largest-remainder rounding): each row $u$ receives $\hat\nu(u) = \mathrm{round}(\nu(u))$ total edges (adjusted to sum to $m{-}k$), distributed across tokens as $c(u,v) = \mathrm{round}(\hat\nu(u) \cdot A_u(v))$ (adjusted to sum to $\hat\nu(u)$).
  \item If an Eulerian path from $s_0$ exists on $c$, find it via Hierholzer's algorithm. Otherwise, try $O(|\cY|^{k+1})$ single-edge swap variants $(c(u,v) \mathrel{-}= 1,\; c(u,v') \mathrel{+}= 1)$ that admit an Eulerian path.
  \item Among all candidates admitting an Eulerian path, select the one maximizing $J_{\mathrm{tilt}}$. The edge labels give tokens $z_{k+1:m}$; prepend the $k$ tokens encoding $s_0$ to recover $z_{1:m}$.
\end{enumerate}

\section{BFTs: detailed derivations}
\label{app:dgp_derivations}

\subsection{Beta--Bernoulli}
\label{app:beta_bernoulli_derivations}

This appendix records the 0-Markov exchangeable (Beta--Bernoulli) specialization of the framework. Write $p:=\tilde p(1)\in[0,1]$.

\paragraph{Posterior.}

The posterior predictive given context $y_{1:n}$ is
\[
  P(Y_{n+1}=1\mid y_{1:n}) = \E[\tilde p\mid y_{1:n}] = \frac{\alpha+S_1(n)}{\alpha+\beta+n},
  \qquad S_1(n) = \sum_{i=1}^n y_i,
\]
and the posterior on $\tilde p$ is $\Pi(\tilde p\mid y_{1:n}) = \mathrm{Beta}\!\big(\alpha+S_1(n),\;\beta+n-S_1(n)\big)$. We use $\alpha=\beta=1/2$.

\paragraph{Reverse cross-entropy: closed form for $J$.}

The reverse cross-entropy utility $U_{\tau^\star}(y_{1:N})=S_1\log\tau^\star + S_0\log(1-\tau^\star)$ is linear in the token counts $S_1=\sum_{t=1}^N y_t$ and $S_0=N-S_1$.
Define the average marginal one-rate $\bar p_N(z_{1:m}):=\frac{1}{N}\sum_{t=1}^N\Pr(Y_{n+t}=1\mid y_{1:n},\,z_{1:m})$. Since $\E[S_1]=N\bar p_N(z_{1:m})$,
\begin{equation}
  J(z_{1:m})=N\Big[\log(1-\tau^\star)+\bar p_N(z_{1:m})\,\log\!\frac{\tau^\star}{1-\tau^\star}\Big].
  \label{eq:J_in_terms_of_pbar}
\end{equation}
This closed form relies on the linearity of $U_{\tau^\star}$ in $S_1$. The frequency match and Dyck utilities lack this linearity, so their objective $J(z_{1:m})$ is computed by enumeration over $\cY^N$; the per-latent $\mu$ defined below is nonetheless available in closed form for all three utilities.

\paragraph{Rao--Blackwellized utilities $\mu(\tilde p)$.}

Under $\tilde p^{\otimes N}$, the continuation tokens $Y_{n+1},\dots,Y_{n+N}$ are i.i.d.\ $\mathrm{Bernoulli}(p)$.

\emph{Reverse cross-entropy ($U_{\tau^\star}$).}
By the same linearity argument,
\begin{equation}
  \mu(\tilde p) = N\Big(\log(1-\tau^\star) + p\,\log\!\frac{\tau^\star}{1-\tau^\star}\Big).
  \label{eq:mu_single_coin_closed}
\end{equation}

\emph{Match target frequency ($U_{q^\star}$).}
The empirical frequency $f=\frac{1}{N}\sum_{t=1}^N Y_{n+t}$ has $\E[f]=p$ and $\Var(f)=p(1-p)/N$ under $\tilde p^{\otimes N}$. Since $\E[(f-q^\star)^2]=\Var(f)+(\E[f]-q^\star)^2$,
\[
  \mu(\tilde p) = -\Big[\frac{p(1-p)}{N} + (p-q^\star)^2\Big].
\]

\emph{Dyck validity ($U_{\mathrm{dyck}}$).}
With $N=4$, the two valid Dyck sequences $\{0101,\,0011\}$ each have probability $p^2(1-p)^2$ under $\tilde p^{\otimes 4}$ (two zeros and two ones in each). Hence
\[
  \mu(\tilde p) = 2\,p^2(1-p)^2.
\]

\subsection{Reinforced urn}
\label{app:reinforced_urn_derivations}

This appendix records the 1-Markov exchangeable (reinforced urn) specialization of the framework, with $k=1$ and $|\cY|=2$.

\paragraph{Predictive and posterior.}

By the Diaconis--Freedman representation theorem \cite{fortiniExchangeabilityPredictionPredictive2025}, the induced predictive rule takes the \emph{reinforced urn} form
\[
  P(Y_{n+1}=y\mid y_{1:n}) = \frac{\alpha/|\cY| + T_{y_n,y}(y_{1:n})}{\alpha + \sum_{y'\in\cY} T_{y_n,y'}(y_{1:n})},
\]
where $T_{a,b}(y_{1:n}) := \sum_{t=1}^{n-1}\mathbf{1}\{y_t=a,\,y_{t+1}=b\}$ counts observed transitions. The posterior on $\tilde Q$ has independent Dirichlet rows
\[
  \Pi(\tilde Q_{a,\cdot}\mid y_{1:n}) = \mathrm{Dir}\!\big(\alpha/|\cY|+T_{a,b}(y_{1:n})\big)_{b\in\cY},\qquad a\in\cY.
\]
We use $\alpha=1$.

\paragraph{Reverse cross-entropy: closed form for $J$.}

The reverse cross-entropy utility $U_{Q^\star}(y_{1:N})=\sum_{t=0}^{N-1}\log Q^\star_{s_t,y_{n+t+1}}$ is a sum of per-step terms. By linearity of expectation,
\[
  J(z_{1:m}) = \sum_{t=0}^{N-1}\sum_{u\in\cY}\sum_{y\in\cY} P(s_t=u,\,Y_{n+t+1}=y\mid y_{1:n},\,z_{1:m})\,\log Q^\star_{u,y},
\]
where the joint marginals $P(s_t=u,\,Y_{n+t+1}=y\mid y_{1:n},\,z_{1:m})$ can be computed from the model's one-step predictives without enumerating all $|\cY|^N$ sequences.
The frequency match and Dyck utilities do not decompose as per-step sums and are computed by enumeration.

\paragraph{Rao--Blackwellized utilities $\mu(\tilde Q;\, s)$.}

Conditional on $\tilde Q$ and starting state $s\in\cY$, the continuation $Y_{1:N}$ is a Markov chain on $\cY$.

\emph{Reverse cross-entropy ($U_{Q^\star}$).}
The utility decomposes over the $N$-step context visitation:
\[
  \mu(\tilde Q;\, s)=\sum_{t=0}^{N-1}\sum_{u\in\cY}\big[e_s^\top \tilde Q^t\big]_u\,h(u,\tilde Q),
\]
where $h(u,\tilde Q):=\sum_{y\in\cY}\tilde Q_{u,y}\log Q^\star_{u,y}$ is the per-state negative cross-entropy and $e_s^\top\tilde Q^t$ is the visitation probability over states at step $t$ (obtained by iterating the $|\cY|\times|\cY|$ transition matrix).

\emph{Match target frequency ($U_{q^\star}$).}
Let $p_t := \big[e_s^\top \tilde Q^t\big]_1$. Then $\E_{\tilde Q}[f]=\frac{1}{N}\sum_{t=1}^N p_t$ and
\[
  \Var_{\tilde Q}(f) = \frac{1}{N^2}\sum_{i,j}\Cov_{\tilde Q}(Y_{i},Y_{j}),
\]
where the covariances follow from the Markov property. Hence
\[
  \mu(\tilde Q;\, s) = -\Big[\Var_{\tilde Q}(f) + \big(\E_{\tilde Q}[f]-q^\star\big)^2\Big].
\]

\emph{Dyck validity ($U_{\mathrm{dyck}}$).}
With $N=4$, the two valid Dyck sequences $\{0101,\,0011\}$ have Markov chain probabilities:
\[
  \mu(\tilde Q;\, s) = \tilde Q_{s,0}\,\tilde Q_{0,1}\,\tilde Q_{1,0}\,\tilde Q_{0,1}
  \;+\;
  \tilde Q_{s,0}\,\tilde Q_{0,0}\,\tilde Q_{0,1}\,\tilde Q_{1,1}.
\]

\subsection{Summary tables}
\label{app:closed_forms}

\begin{table}[h]
  \centering
  \caption{Exact objective $J(z_{1:m})$ for each BFT/utility pair. We write $\bar p_N(z_{1:m}) := \frac{1}{N}\sum_{t=1}^N \Pr(Y_t=1\mid z_{1:m})$ for the average marginal one-rate.}
  \label{tab:J_spelled_out}
  \renewcommand{\arraystretch}{2.2}
  \setlength{\tabcolsep}{6pt}
  \begin{tabular}{l l l}
    \toprule
    Utility & Beta--Bernoulli                                                                            & Reinforced urn \\
    \midrule
    Rev.\ xent.
            & $N\!\Big[\log(1\!-\!\tau^\star)+\bar p_N(z_{1:m})\log\frac{\tau^\star}{1-\tau^\star}\Big]$
            & $\displaystyle\sum_{t,u,y} P(s_t\!=\!u, Y_{t+1}\!=\!y)\log Q^\star_{u,y}$                                   \\
    Freq.\ match
            & enumerate
            & enumerate                                                                                                   \\
    Dyck
            & $\pi_\theta(\{0101,0011\}\mid z_{1:m})$
            & $\pi_\theta(\{0101,0011\}\mid z_{1:m})$                                                                     \\
    \bottomrule
  \end{tabular}
\end{table}

\begin{table}[h]
  \centering
  \caption{Closed-form Rao--Blackwellized utility $\mu$ for each setting. Here $p := \tilde p(1)$, $h(u, \tilde Q) := \sum_y \tilde Q_{u,y}\log Q^\star_{u,y}$, and $P(s_t = u)$ denotes the context-visitation probability under $\tilde Q$ starting from state $s$.}
  \label{tab:rb_mu}
  \renewcommand{\arraystretch}{2.2}
  \setlength{\tabcolsep}{6pt}
  \begin{tabular}{l l l}
    \toprule
    Utility & Beta--Bernoulli: $\mu(\tilde p)$                                                                                      & Reinforced urn: $\mu(\tilde Q;\, s)$ \\
    \midrule
    Rev.\ xent.
            & $N\!\Big(\log(1\!-\!\tau^\star)+p\log\frac{\tau^\star}{1-\tau^\star}\Big)$
            & $\displaystyle\sum_{t=0}^{N-1}\sum_u P(s_t\!=\!u)\,h(u, \tilde Q)$                                                                                           \\
    Freq.\ match
            & $-\!\Big[\dfrac{p(1-p)}{N}+(p-q^\star)^2\Big]$
            & $-\!\Big[\Var_{\tilde Q}(f) + (\E_{\tilde Q}[f]-q^\star)^2\Big]$                                                                                             \\
    Dyck
            & $2\,p^2(1-p)^2$
            & $\tilde Q_{s,0}\tilde Q_{0,1}\tilde Q_{1,0}\tilde Q_{0,1} + \tilde Q_{s,0}\tilde Q_{0,0}\tilde Q_{0,1}\tilde Q_{1,1}$                                        \\
    \bottomrule
  \end{tabular}
\end{table}

The reverse cross-entropy admits a closed-form $J$ for both BFTs because $U$ is a sum of per-step terms; the other utilities require enumeration over $\cY^N$. All three $\mu$ closed forms hold in both BFTs: Beta--Bernoulli benefits from i.i.d.\ continuation tokens; the reinforced urn uses Markov-chain marginals via $\tilde Q^t$ for the reverse cross-entropy and frequency-match utilities, and direct enumeration of the two valid Dyck sequences for the Dyck utility.

\section{Experiments details}
\label{app:experiments}

\subsection{Utility functions}
\label{app:utility_specs}

The reverse cross-entropy utility is fully specified in Section~\ref{sec:utilities}. We give the corresponding details for the frequency-match and Dyck utilities here.

\paragraph{Match target frequency.} Fix a target frequency $q^\star\in[0,1]$ and let $f(y_{1:N}):=\frac{1}{N}\sum_{t=1}^N y_t$ be the empirical frequency of token $1$ in the continuation. Set $U_{q^\star}(y_{1:N}) := -\big(f(y_{1:N}) - q^\star\big)^2$, the negative squared deviation of the continuation frequency from the target. Maximizing $J$ pulls the continuation frequency toward $q^\star$ in mean square; the utility depends on $y_{1:N}$ only through its count of $1$s. We sweep $q^\star\in\{0,\,0.1,\,0.2,\,\dots,\,1.0\}$ (11 targets).

\paragraph{Dyck validity.} Identify $0$ with ``\texttt{(}'' and $1$ with ``\texttt{)}''. A sequence $y_{1:N}\in\cY^N$ is a \emph{valid Dyck} sequence if every prefix contains at least as many ``\texttt{(}'' as ``\texttt{)}'' and the full sequence is balanced. We set $U_{\mathrm{dyck}}(y_{1:N}) = 1$ when $y_{1:N}$ is a valid Dyck sequence and $0$ otherwise. With $N=4$ the valid set is $\{0101,\,0011\}$ out of $|\cY|^N=16$ continuations, so $J = \E\,[U_{\mathrm{dyck}}] = \pi_\theta(\text{valid Dyck}\mid z_{1:m})$.

\subsection{BFT training}
\label{app:bft_training}

\paragraph{Beta--Bernoulli transformer.}
\label{app:bb_architecture}
1-layer pre-norm transformer with $d_{\mathrm{model}}=64$, 4 attention heads ($d_{\mathrm{head}}=16$), and feedforward dimension $128$. Token embeddings $\phi\colon\cY\cup\{\mathrm{BOS}\}\to\mathbb{R}^{64}$ are learned (3 embeddings: tokens 0, 1, and BOS); no positional encoding is used. Every training and inference sequence is prepended with a BOS token. Each training batch consists of $128$ sequences of length $2000$, each generated by sampling $\tilde p\sim\mathrm{Beta}(1/2,1/2)$ and generating an i.i.d.\ $\mathrm{Bernoulli}(\tilde p)$ sequence. The model is trained for $50{,}000$ iterations with AdamW ($\beta_1=0.9$, $\beta_2=0.95$, weight decay $0.1$), cosine learning rate schedule from $3\times 10^{-4}$ to $10^{-5}$ with $500$ warmup steps, bfloat16 mixed precision, and gradient clipping at norm $1.0$.

\paragraph{Reinforced urn transformer.}
\label{app:ru_architecture}
8-layer pre-norm transformer with $d_{\mathrm{model}}=256$, 8 attention heads ($d_{\mathrm{head}}=32$), and feedforward dimension $1024$. Token embeddings $\phi\colon\cY\to\mathbb{R}^{256}$ and position embeddings $\psi\colon\{0,\ldots,19999\}\to\mathbb{R}^{256}$ are both learned; at each position $t$ the input is $\phi(y_t)+\psi(t)$. A BOS token is prepended, with its own learned embedding. Each training batch consists of $64$ sequences of length $2000$, each generated by sampling $\tilde Q_{u,\cdot}\overset{\mathrm{i.i.d.}}{\sim}\mathrm{Dir}(1/2,1/2)$ for $u\in\{0,1\}$ and generating a 1-Markov chain from $\tilde Q$ with a random initial token. The model is trained for $100{,}000$ iterations with AdamW ($\beta_1=0.9$, $\beta_2=0.95$, weight decay $0.1$), cosine learning rate schedule from $3\times 10^{-4}$ to $10^{-5}$ with $1{,}000$ warmup steps, bfloat16 mixed precision, and gradient clipping at norm $1.0$.

\subsection{PPT and PPT-RB}
\label{app:ppt_details}

The four PPT variants share the parameterization, optimizer, and IS gradient estimator below; they differ in (a) how prior samples are obtained and (b) how $\widehat\mu^{(\ell)}$ ($k=0$) or $\widehat{\bar\mu}^{(\ell)}$ ($k\geq 1$) in \eqref{eq:IS_estimator} or \eqref{eq:IS_estimator_kmarkov} is computed.

\paragraph{Parameterization and optimizer.} For $k=0$, $\alpha\in\Delta^1$ is parameterized by unconstrained logits $(a_0, a_1)\in\mathbb{R}^2$ via softmax. For $k\geq 1$, $A$ and $\rho$ are optimized jointly by simplex-projected gradient descent with learning rate $\eta = 0.1$. For $A$, each row $A_{s,\cdot}$ is updated independently:
\[
  A_{s,\cdot} \leftarrow \mathrm{proj}_\Delta\!\Big(A_{s,\cdot} - \eta\,\big(\nabla_{A_{s,\cdot}} J_{\mathrm{tilt}} - \overline{\nabla_{A_{s,\cdot}} J_{\mathrm{tilt}}}\big)\Big),
\]
where $\overline{(\cdot)}$ denotes the mean over the $|\cY|$ components and $\mathrm{proj}_\Delta(x)_v := \max(x_v,\,\epsilon)\big/\sum_{v'}\max(x_{v'},\,\epsilon)$ with $\epsilon=10^{-6}$; $\rho$ is updated identically (Appendix~\ref{app:rho_update}). The gradient $\nabla_A J_{\mathrm{tilt}}$ is computed by differentiating through the matrix power of $M^{(\tilde Q)}$ (Appendix~\ref{app:W_closed_form}). Both $A$ and $\rho$ are initialized by independent draws from $\mathrm{Dir}(1,\dots,1)$. Optimization terminates early if $J_{\mathrm{tilt}}$ does not improve by more than $10^{-5}$ for $100$ consecutive steps.

\paragraph{Drawing prior samples.} For \textsc{PPT} and \textsc{PPT-RB}, $L=5{,}000$ samples are obtained from the BFT's latent prior by Predictive Monte Carlo (Appendix~\ref{app:pmc}) using rollouts of length $R=2{,}000$ from the trained BFT $\pi_\theta$. For the \textsc{(analytic)} variants, $L=5{,}000$ samples are drawn directly from $\Pi_0$: $\tilde p\sim\mathrm{Beta}(1/2,1/2)$ for the Beta--Bernoulli BFT and rows $\tilde Q_{a,\cdot}\stackrel{\text{i.i.d.}}{\sim}\mathrm{Dir}(1/2,1/2)$ for the reinforced urn BFT. In both cases the $L$ samples are drawn once at the start and reused across optimization steps.

\paragraph{Runtime.} The full experiment sweep completes in approximately $8$ hours on a single NVIDIA H100 GPU.

\paragraph{Computing $\widehat\mu^{(\ell)}$ and $\widehat{\bar\mu}^{(\ell)}$.} \textsc{PPT-RB} uses the closed-form $\mu(\tilde p^{(\ell)})$ ($k=0$) or $\bar\mu(\tilde Q^{(\ell)};\, A, \rho)$ ($k\geq 1$) from Appendix~\ref{app:closed_forms}. \textsc{PPT} uses a single rollout per prior sample: for $k=0$ we draw $Y^{(\ell)}_{1:N}\sim (\tilde p^{(\ell)})^{\otimes N}$ and set $\widehat\mu^{(\ell)} = U(Y^{(\ell)}_{1:N})$; for $k\geq 1$ we draw a prompt $z^{(\ell)}\sim P_{A,\rho}$, then a continuation $Y^{(\ell)}_{1:N}\sim \tilde Q^{(\ell)}(\cdot\mid \mathrm{suf}_k(z^{(\ell)}))$, and set $\widehat{\bar\mu}^{(\ell)} = U(Y^{(\ell)}_{1:N})$.

\subsection{GCG}
\label{app:gcg_details}
Each iteration, if $z_{1:m}^{(t)}$ is the prompt from iteration $t$, the GCG algorithm evaluates the gradient
\begin{equation}
  \nabla_{e_{z_i^{(t)}}} J(z_{1:m}^{(t)})\in \mathbb R^{|\mathcal Y|}
\end{equation}

where $e_{z_i^{(t)}}$ denotes the one-hot vector representing the value of the $i$th token, and uses this to select the top-$k$ most promising token substitutions $\mathcal X_i = \text{Top-}k(\nabla_{e_{z_i^{(t)}}} J(z_{1:m}^{(t)}))$ for each $i \in \{1, \dots, m\}$.

In our binary setting $|\mathcal Y|=2$, so any $k\geq 2$ leaves $\mathcal X_i = \{0,1\}$ and the gradient ranking has no effect. We therefore set $k=1$, so $\mathcal X_i = \{c_i^{(t)}\}$ where $c_i^{(t)} \in \arg\max_v \nabla_{e_{z_i^{(t)},v}} J(z_{1:m}^{(t)})$ is the gradient's Top-1 choice at position $i$. Let $S^{(t)} = \{i : c_i^{(t)} \neq z_i^{(t)}\}$ be the positions where the Top-1 token differs from the current one, and let $z^{(t,i)}$ denote $z_{1:m}^{(t)}$ with position $i$ replaced by $c_i^{(t)}$. The candidate set at iteration $t$ is $\{z^{(t,i)} : i \in S^{(t)}\}$; we update by
$$
  z_{1:m}^{(t+1)} = \arg\max_{i \in S^{(t)}} J(z^{(t,i)}),
$$
setting $z_{1:m}^{(t+1)} = z_{1:m}^{(t)}$ when $S^{(t)} = \emptyset$ or when no candidate improves $J$. Initialization is $z_i^{(0)} \stackrel{\mathrm{i.i.d.}}{\sim} \mathrm{Uniform}(\{0,1\})$. Each iteration costs one backprop through $\pi_\theta$ to obtain the gradient, plus $|S^{(t)}|\cdot|\mathcal Y|^N$ forwards (at most $m\cdot|\mathcal Y|^N$) to evaluate the candidates. We run $\max(10,\,2m)$ iterations (12 for $m=6$, 100 for $m=50$), halting early when no candidate improves $J$.

\section{Experimental results}
\label{app:experiments-results}

\subsection{Method comparison}
\label{app:susan-summary}

This subsection reports per-configuration performance for the five methods, averaged over 10 random seeds. The convention is that $J$ is to be \emph{maximised}; for $m=6$, where the full set of $V^m=64$ hard prompts is enumerable, we additionally report rank-out-of-$64$ of the snapped prompt (rank $1$ = best). Each cell averages $J$ (or rank) over $n=10$ random seeds, with standard error in parentheses; the best method per row is in bold (within $10^{-3}$ for $J$, within $0.05$ for rank). Reverse cross-entropy on the reinforced urn at both $m\in\{6,50\}$ is the main-text Table~\ref{tab:rev-xent-urn}; the corresponding rank table at $m=6$ is Table~\ref{tab:rev-xent-urn-rank}. Reverse cross-entropy on the Beta--Bernoulli BFT is in Tables~\ref{tab:rev-xent-bb} and~\ref{tab:rev-xent-bb-rank}.

\begin{table}[h]
  \centering
  \caption{Reverse cross-entropy on the Beta--Bernoulli BFT. Each cell is the mean of $J$ over $n=10$ random seeds (max convention; higher better), with standard error in parentheses. Best method per $m$ in bold (within $10^{-3}$).}
  \label{tab:rev-xent-bb}
  \resizebox{\textwidth}{!}{%
    \footnotesize
    \begin{tabular}{l rrrrr rrrrr}
      \toprule
                       & \multicolumn{5}{c}{$m=6$} & \multicolumn{5}{c}{$m=50$}                                                                                                                                                                                                       \\
      \cmidrule(lr){2-6} \cmidrule(lr){7-11}
                       & GCG                       & PPT-RB (A)                 & PPT-RB                   & PPT (A)                  & PPT                      & GCG                      & PPT-RB (A)               & PPT-RB                   & PPT (A)         & PPT             \\
      \midrule
      $\tau^\star=0.1$ & $\mathbf{-1.06}\,(0.00)$  & $\mathbf{-1.06}\,(0.00)$   & $\mathbf{-1.06}\,(0.00)$ & $\mathbf{-1.06}\,(0.00)$ & $\mathbf{-1.06}\,(0.00)$ & $\mathbf{-0.51}\,(0.00)$ & $\mathbf{-0.51}\,(0.00)$ & $\mathbf{-0.51}\,(0.00)$ & $-1.16\,(0.18)$ & $-1.23\,(0.20)$ \\
      $\tau^\star=0.2$ & $\mathbf{-1.29}\,(0.00)$  & $\mathbf{-1.29}\,(0.00)$   & $\mathbf{-1.29}\,(0.00)$ & $\mathbf{-1.29}\,(0.00)$ & $\mathbf{-1.29}\,(0.00)$ & $\mathbf{-0.95}\,(0.00)$ & $\mathbf{-0.95}\,(0.00)$ & $\mathbf{-0.95}\,(0.00)$ & $-1.36\,(0.12)$ & $-1.40\,(0.13)$ \\
      $\tau^\star=0.3$ & $\mathbf{-1.67}\,(0.00)$  & $\mathbf{-1.67}\,(0.00)$   & $\mathbf{-1.67}\,(0.00)$ & $\mathbf{-1.67}\,(0.00)$ & $\mathbf{-1.67}\,(0.00)$ & $\mathbf{-1.46}\,(0.00)$ & $\mathbf{-1.46}\,(0.00)$ & $\mathbf{-1.46}\,(0.00)$ & $-1.71\,(0.07)$ & $-1.74\,(0.08)$ \\
      $\tau^\star=0.4$ & $\mathbf{-2.16}\,(0.00)$  & $\mathbf{-2.16}\,(0.00)$   & $\mathbf{-2.16}\,(0.00)$ & $\mathbf{-2.16}\,(0.00)$ & $\mathbf{-2.16}\,(0.00)$ & $\mathbf{-2.06}\,(0.00)$ & $\mathbf{-2.06}\,(0.00)$ & $\mathbf{-2.06}\,(0.00)$ & $-2.18\,(0.03)$ & $-2.19\,(0.04)$ \\
      $\tau^\star=0.6$ & $\mathbf{-2.16}\,(0.00)$  & $\mathbf{-2.16}\,(0.00)$   & $\mathbf{-2.16}\,(0.00)$ & $\mathbf{-2.16}\,(0.00)$ & $\mathbf{-2.16}\,(0.00)$ & $\mathbf{-2.06}\,(0.00)$ & $\mathbf{-2.06}\,(0.00)$ & $\mathbf{-2.06}\,(0.00)$ & $-2.17\,(0.03)$ & $-2.18\,(0.04)$ \\
      $\tau^\star=0.7$ & $\mathbf{-1.67}\,(0.00)$  & $\mathbf{-1.67}\,(0.00)$   & $\mathbf{-1.67}\,(0.00)$ & $\mathbf{-1.67}\,(0.00)$ & $\mathbf{-1.67}\,(0.00)$ & $\mathbf{-1.46}\,(0.00)$ & $\mathbf{-1.46}\,(0.00)$ & $\mathbf{-1.46}\,(0.00)$ & $-1.69\,(0.07)$ & $-1.71\,(0.08)$ \\
      $\tau^\star=0.8$ & $\mathbf{-1.29}\,(0.00)$  & $\mathbf{-1.29}\,(0.00)$   & $\mathbf{-1.29}\,(0.00)$ & $\mathbf{-1.29}\,(0.00)$ & $\mathbf{-1.29}\,(0.00)$ & $\mathbf{-0.95}\,(0.00)$ & $\mathbf{-0.95}\,(0.00)$ & $\mathbf{-0.95}\,(0.00)$ & $-1.32\,(0.12)$ & $-1.36\,(0.13)$ \\
      $\tau^\star=0.9$ & $\mathbf{-1.05}\,(0.00)$  & $\mathbf{-1.05}\,(0.00)$   & $\mathbf{-1.05}\,(0.00)$ & $\mathbf{-1.05}\,(0.00)$ & $\mathbf{-1.05}\,(0.00)$ & $\mathbf{-0.51}\,(0.00)$ & $\mathbf{-0.51}\,(0.00)$ & $\mathbf{-0.51}\,(0.00)$ & $-1.10\,(0.19)$ & $-1.17\,(0.21)$ \\
      \bottomrule
    \end{tabular}}
\end{table}

\begin{table}[h]
  \centering
  \caption{Reverse cross-entropy, Beta--Bernoulli, $m=6$. Rank-out-of-64 of the snapped prompt (1 = enumerated optimum), averaged over $n=10$ random seeds; standard error in parentheses. Best method per row in bold (within $0.05$).}
  \label{tab:rev-xent-bb-rank}
  \resizebox{0.85\textwidth}{!}{%
    \begin{tabular}{l rrrrr}
      \toprule
                       & GCG                    & PPT-RB (A)             & PPT-RB                 & PPT (A)                & PPT                    \\
      \midrule
      $\tau^\star=0.1$ & $\mathbf{1.0}\,(0.00)$ & $\mathbf{1.0}\,(0.00)$ & $\mathbf{1.0}\,(0.00)$ & $\mathbf{1.0}\,(0.00)$ & $\mathbf{1.0}\,(0.00)$ \\
      $\tau^\star=0.2$ & $\mathbf{1.0}\,(0.00)$ & $\mathbf{1.0}\,(0.00)$ & $\mathbf{1.0}\,(0.00)$ & $\mathbf{1.0}\,(0.00)$ & $\mathbf{1.0}\,(0.00)$ \\
      $\tau^\star=0.3$ & $\mathbf{1.0}\,(0.00)$ & $\mathbf{1.0}\,(0.00)$ & $\mathbf{1.0}\,(0.00)$ & $\mathbf{1.0}\,(0.00)$ & $\mathbf{1.0}\,(0.00)$ \\
      $\tau^\star=0.4$ & $\mathbf{1.0}\,(0.00)$ & $\mathbf{1.0}\,(0.00)$ & $\mathbf{1.0}\,(0.00)$ & $\mathbf{1.0}\,(0.00)$ & $\mathbf{1.0}\,(0.00)$ \\
      $\tau^\star=0.6$ & $\mathbf{1.0}\,(0.00)$ & $\mathbf{1.0}\,(0.00)$ & $\mathbf{1.0}\,(0.00)$ & $\mathbf{1.0}\,(0.00)$ & $\mathbf{1.0}\,(0.00)$ \\
      $\tau^\star=0.7$ & $\mathbf{1.0}\,(0.00)$ & $\mathbf{1.0}\,(0.00)$ & $\mathbf{1.0}\,(0.00)$ & $\mathbf{1.0}\,(0.00)$ & $\mathbf{1.0}\,(0.00)$ \\
      $\tau^\star=0.8$ & $\mathbf{1.0}\,(0.00)$ & $\mathbf{1.0}\,(0.00)$ & $\mathbf{1.0}\,(0.00)$ & $\mathbf{1.0}\,(0.00)$ & $\mathbf{1.0}\,(0.00)$ \\
      $\tau^\star=0.9$ & $\mathbf{1.0}\,(0.00)$ & $\mathbf{1.0}\,(0.00)$ & $\mathbf{1.0}\,(0.00)$ & $\mathbf{1.0}\,(0.00)$ & $\mathbf{1.0}\,(0.00)$ \\
      \bottomrule
    \end{tabular}}
\end{table}

\begin{table}[h]
  \centering
  \caption{Reverse cross-entropy on the reinforced urn BFT, $m=6$. Rank-out-of-64 of the snapped prompt (1 = enumerated optimum), averaged over $n=10$ random seeds; standard error in parentheses. Best method per row in bold (within $0.05$).}
  \label{tab:rev-xent-urn-rank}
  \resizebox{0.85\textwidth}{!}{%
    \begin{tabular}{l rrrrr}
      \toprule
                       & GCG            & PPT-RB (A)             & PPT-RB                 & PPT (A)                & PPT                    \\
      \midrule
      \texttt{sym-0.1} & $4.9\,(1.76)$  & $\mathbf{1.4}\,(0.16)$ & $2.0\,(0.00)$          & $1.5\,(0.17)$          & $2.0\,(0.00)$          \\
      \texttt{sym-0.2} & $4.9\,(1.76)$  & $\mathbf{1.4}\,(0.16)$ & $2.0\,(0.00)$          & $\mathbf{1.4}\,(0.16)$ & $2.0\,(0.00)$          \\
      \texttt{sym-0.3} & $4.9\,(1.76)$  & $\mathbf{1.5}\,(0.17)$ & $2.0\,(0.00)$          & $1.6\,(0.16)$          & $2.0\,(0.00)$          \\
      \texttt{sym-0.4} & $4.9\,(1.76)$  & $\mathbf{1.5}\,(0.17)$ & $2.0\,(0.00)$          & $1.6\,(0.16)$          & $2.0\,(0.00)$          \\
      \texttt{sym-0.6} & $11.0\,(3.52)$ & $\mathbf{1.3}\,(0.15)$ & $\mathbf{1.3}\,(0.15)$ & $\mathbf{1.3}\,(0.15)$ & $\mathbf{1.3}\,(0.15)$ \\
      \texttt{sym-0.7} & $11.0\,(3.52)$ & $\mathbf{1.3}\,(0.15)$ & $\mathbf{1.3}\,(0.15)$ & $\mathbf{1.3}\,(0.15)$ & $\mathbf{1.3}\,(0.15)$ \\
      \texttt{sym-0.8} & $11.0\,(3.52)$ & $\mathbf{1.3}\,(0.15)$ & $\mathbf{1.3}\,(0.15)$ & $\mathbf{1.3}\,(0.15)$ & $\mathbf{1.3}\,(0.15)$ \\
      \texttt{sym-0.9} & $11.0\,(3.52)$ & $\mathbf{1.3}\,(0.15)$ & $\mathbf{1.3}\,(0.15)$ & $\mathbf{1.3}\,(0.15)$ & $\mathbf{1.3}\,(0.15)$ \\
      \texttt{sym-1.0} & $11.0\,(3.52)$ & $\mathbf{1.5}\,(0.17)$ & $1.6\,(0.16)$          & $\mathbf{1.5}\,(0.17)$ & $1.6\,(0.16)$          \\
      \midrule
      \texttt{dir-0}   & $16.1\,(3.90)$ & $\mathbf{1.0}\,(0.00)$ & $2.4\,(1.40)$          & $\mathbf{1.0}\,(0.00)$ & $2.4\,(1.40)$          \\
      \texttt{dir-1}   & $12.7\,(3.68)$ & $\mathbf{1.0}\,(0.00)$ & $\mathbf{1.0}\,(0.00)$ & $\mathbf{1.0}\,(0.00)$ & $\mathbf{1.0}\,(0.00)$ \\
      \texttt{dir-2}   & $9.3\,(2.68)$  & $\mathbf{1.0}\,(0.00)$ & $\mathbf{1.0}\,(0.00)$ & $\mathbf{1.0}\,(0.00)$ & $\mathbf{1.0}\,(0.00)$ \\
      \texttt{dir-3}   & $10.5\,(3.33)$ & $\mathbf{1.6}\,(0.31)$ & $\mathbf{1.6}\,(0.31)$ & $\mathbf{1.6}\,(0.31)$ & $\mathbf{1.6}\,(0.31)$ \\
      \texttt{dir-4}   & $7.0\,(1.33)$  & $2.9\,(0.82)$          & $4.4\,(2.57)$          & $2.9\,(0.82)$          & $\mathbf{2.8}\,(0.76)$ \\
      \texttt{dir-5}   & $5.0\,(1.59)$  & $\mathbf{1.0}\,(0.00)$ & $\mathbf{1.0}\,(0.00)$ & $\mathbf{1.0}\,(0.00)$ & $\mathbf{1.0}\,(0.00)$ \\
      \texttt{dir-6}   & $10.9\,(3.07)$ & $3.0\,(0.82)$          & $\mathbf{2.5}\,(0.76)$ & $3.0\,(0.82)$          & $\mathbf{2.5}\,(0.76)$ \\
      \texttt{dir-7}   & $11.3\,(3.54)$ & $\mathbf{1.6}\,(0.16)$ & $\mathbf{1.6}\,(0.16)$ & $1.7\,(0.15)$          & $\mathbf{1.6}\,(0.16)$ \\
      \texttt{dir-8}   & $10.8\,(5.12)$ & $\mathbf{5.6}\,(2.40)$ & $\mathbf{5.6}\,(2.40)$ & $\mathbf{5.6}\,(2.40)$ & $\mathbf{5.6}\,(2.40)$ \\
      \texttt{dir-9}   & $13.5\,(4.36)$ & $\mathbf{1.0}\,(0.00)$ & $3.5\,(2.50)$          & $\mathbf{1.0}\,(0.00)$ & $3.5\,(2.50)$          \\
      \bottomrule
    \end{tabular}}
\end{table}

\paragraph{Frequency match.} On Beta--Bernoulli at $m=6$ (Tables~\ref{tab:freq-match-summary} and \ref{tab:freq-match-rank}), all methods reach rank 1 for most $q^\star$ values. At $q^\star \in \{0.6, 0.7\}$, the \textsc{PPT} variants land on suboptimal prompts (rank 6--11) while \textsc{GCG} reaches rank 1--4; at $q^\star \in \{0.3, 0.4\}$ the situation reverses. On the reinforced urn at $m=6$, the \textsc{PPT} variants substantially outperform \textsc{GCG} at the extremes ($q^\star \in \{0.0, 0.1, 0.2, 0.8, 0.9, 1.0\}$: \textsc{PPT} reaches rank 1--4 while \textsc{GCG} ranks 15--18 of 64), are roughly tied for $q^\star \in \{0.4, 0.5, 0.6\}$, and all methods rank substantially below the optimum at $q^\star = 0.7$. At $m=50$ on Beta--Bernoulli, \textsc{GCG} achieves $J$ closer to zero than the \textsc{PPT} variants in every cell; on the reinforced urn at $m=50$, \textsc{GCG} and \textsc{PPT-RB} are roughly competitive, with \textsc{PPT} (without Rao--Blackwellisation) consistently lagging.

\begin{table}[h]
  \centering
  \caption{Frequency match. Each cell is the mean of $J$ over $n=10$ random seeds (max convention; higher better), with standard error in parentheses. Best method per ($k$, $m$) in bold (within $10^{-3}$).}
  \label{tab:freq-match-summary}
  \resizebox{\textwidth}{!}{%
    \footnotesize
    \begin{tabular}{l rrrrr rrrrr}
      \toprule
                    & \multicolumn{5}{c}{$m=6$} & \multicolumn{5}{c}{$m=50$}                                                                                                                                                                                                                         \\
      \cmidrule(lr){2-6} \cmidrule(lr){7-11}
                    & GCG                       & PPT-RB (A)                 & PPT-RB                   & PPT (A)                  & PPT                      & GCG                      & PPT-RB (A)               & PPT-RB                   & PPT (A)                  & PPT                      \\
      \midrule
      \multicolumn{11}{l}{\emph{Beta--Bernoulli ($k=0$)}}                                                                                                                                                                                                                                            \\
      $q^\star=0.0$ & $\mathbf{-0.03}\,(0.00)$  & $\mathbf{-0.03}\,(0.00)$   & $\mathbf{-0.03}\,(0.00)$ & $\mathbf{-0.03}\,(0.00)$ & $\mathbf{-0.03}\,(0.00)$ & $\mathbf{-0.00}\,(0.00)$ & $-0.01\,(0.00)$          & $-0.01\,(0.00)$          & $-0.04\,(0.01)$          & $-0.04\,(0.01)$          \\
      $q^\star=0.1$ & $\mathbf{-0.02}\,(0.00)$  & $\mathbf{-0.02}\,(0.00)$   & $\mathbf{-0.02}\,(0.00)$ & $\mathbf{-0.02}\,(0.00)$ & $\mathbf{-0.02}\,(0.00)$ & $\mathbf{-0.01}\,(0.00)$ & $-0.02\,(0.00)$          & $-0.02\,(0.00)$          & $-0.03\,(0.01)$          & $-0.03\,(0.01)$          \\
      $q^\star=0.2$ & $\mathbf{-0.04}\,(0.00)$  & $\mathbf{-0.04}\,(0.00)$   & $\mathbf{-0.04}\,(0.00)$ & $\mathbf{-0.04}\,(0.00)$ & $\mathbf{-0.04}\,(0.00)$ & $\mathbf{-0.03}\,(0.00)$ & $-0.07\,(0.01)$          & $-0.07\,(0.01)$          & $-0.07\,(0.01)$          & $-0.07\,(0.01)$          \\
      $q^\star=0.3$ & $\mathbf{-0.07}\,(0.00)$  & $\mathbf{-0.07}\,(0.00)$   & $\mathbf{-0.07}\,(0.00)$ & $\mathbf{-0.07}\,(0.00)$ & $\mathbf{-0.07}\,(0.00)$ & $\mathbf{-0.05}\,(0.00)$ & $-0.06\,(0.01)$          & $-0.06\,(0.01)$          & $-0.06\,(0.01)$          & $-0.06\,(0.01)$          \\
      $q^\star=0.4$ & $\mathbf{-0.08}\,(0.00)$  & $\mathbf{-0.08}\,(0.00)$   & $\mathbf{-0.08}\,(0.00)$ & $\mathbf{-0.08}\,(0.00)$ & $\mathbf{-0.08}\,(0.00)$ & $\mathbf{-0.06}\,(0.00)$ & $-0.07\,(0.00)$          & $-0.07\,(0.00)$          & $-0.07\,(0.00)$          & $-0.07\,(0.00)$          \\
      $q^\star=0.5$ & $\mathbf{-0.09}\,(0.00)$  & $\mathbf{-0.09}\,(0.00)$   & $\mathbf{-0.09}\,(0.00)$ & $\mathbf{-0.09}\,(0.00)$ & $\mathbf{-0.09}\,(0.00)$ & $\mathbf{-0.07}\,(0.00)$ & $\mathbf{-0.07}\,(0.00)$ & $\mathbf{-0.07}\,(0.00)$ & $\mathbf{-0.07}\,(0.00)$ & $\mathbf{-0.07}\,(0.00)$ \\
      $q^\star=0.6$ & $\mathbf{-0.08}\,(0.00)$  & $\mathbf{-0.08}\,(0.00)$   & $\mathbf{-0.08}\,(0.00)$ & $\mathbf{-0.08}\,(0.00)$ & $\mathbf{-0.08}\,(0.00)$ & $\mathbf{-0.06}\,(0.00)$ & $-0.07\,(0.00)$          & $-0.07\,(0.00)$          & $-0.07\,(0.00)$          & $-0.07\,(0.00)$          \\
      $q^\star=0.7$ & $\mathbf{-0.07}\,(0.00)$  & $\mathbf{-0.07}\,(0.00)$   & $\mathbf{-0.07}\,(0.00)$ & $\mathbf{-0.07}\,(0.00)$ & $\mathbf{-0.07}\,(0.00)$ & $\mathbf{-0.05}\,(0.00)$ & $-0.07\,(0.01)$          & $-0.07\,(0.01)$          & $-0.07\,(0.01)$          & $-0.06\,(0.01)$          \\
      $q^\star=0.8$ & $\mathbf{-0.04}\,(0.00)$  & $\mathbf{-0.04}\,(0.00)$   & $\mathbf{-0.04}\,(0.00)$ & $\mathbf{-0.04}\,(0.00)$ & $\mathbf{-0.04}\,(0.00)$ & $\mathbf{-0.03}\,(0.00)$ & $-0.08\,(0.01)$          & $-0.08\,(0.01)$          & $-0.08\,(0.01)$          & $-0.08\,(0.01)$          \\
      $q^\star=0.9$ & $\mathbf{-0.02}\,(0.00)$  & $\mathbf{-0.02}\,(0.00)$   & $\mathbf{-0.02}\,(0.00)$ & $\mathbf{-0.02}\,(0.00)$ & $\mathbf{-0.02}\,(0.00)$ & $\mathbf{-0.01}\,(0.00)$ & $-0.02\,(0.00)$          & $-0.02\,(0.00)$          & $-0.03\,(0.01)$          & $-0.03\,(0.01)$          \\
      $q^\star=1.0$ & $\mathbf{-0.03}\,(0.00)$  & $\mathbf{-0.03}\,(0.00)$   & $\mathbf{-0.03}\,(0.00)$ & $\mathbf{-0.03}\,(0.00)$ & $\mathbf{-0.03}\,(0.00)$ & $\mathbf{-0.00}\,(0.00)$ & $-0.01\,(0.00)$          & $-0.01\,(0.00)$          & $-0.04\,(0.01)$          & $-0.04\,(0.01)$          \\
      \midrule
      \multicolumn{11}{l}{\emph{Reinforced urn ($k=1$)}}                                                                                                                                                                                                                                             \\
      $q^\star=0.0$ & $-0.19\,(0.04)$           & $\mathbf{-0.07}\,(0.02)$   & $\mathbf{-0.07}\,(0.02)$ & $\mathbf{-0.07}\,(0.02)$ & $\mathbf{-0.07}\,(0.02)$ & $-0.03\,(0.00)$          & $\mathbf{-0.00}\,(0.00)$ & $-0.01\,(0.01)$          & $-0.06\,(0.04)$          & $-0.08\,(0.04)$          \\
      $q^\star=0.1$ & $-0.13\,(0.03)$           & $\mathbf{-0.05}\,(0.02)$   & $\mathbf{-0.05}\,(0.02)$ & $\mathbf{-0.05}\,(0.02)$ & $\mathbf{-0.05}\,(0.02)$ & $\mathbf{-0.02}\,(0.00)$ & $-0.02\,(0.01)$          & $-0.03\,(0.01)$          & $-0.06\,(0.02)$          & $-0.06\,(0.02)$          \\
      $q^\star=0.2$ & $-0.10\,(0.02)$           & $\mathbf{-0.05}\,(0.01)$   & $\mathbf{-0.05}\,(0.01)$ & $\mathbf{-0.05}\,(0.01)$ & $\mathbf{-0.05}\,(0.01)$ & $\mathbf{-0.03}\,(0.00)$ & $-0.05\,(0.01)$          & $-0.05\,(0.01)$          & $-0.06\,(0.01)$          & $-0.07\,(0.01)$          \\
      $q^\star=0.3$ & $-0.07\,(0.01)$           & $\mathbf{-0.06}\,(0.00)$   & $-0.07\,(0.00)$          & $\mathbf{-0.06}\,(0.00)$ & $-0.07\,(0.00)$          & $\mathbf{-0.03}\,(0.00)$ & $-0.05\,(0.01)$          & $-0.05\,(0.01)$          & $-0.06\,(0.01)$          & $-0.06\,(0.01)$          \\
      $q^\star=0.4$ & $\mathbf{-0.03}\,(0.00)$  & $\mathbf{-0.03}\,(0.00)$   & $\mathbf{-0.03}\,(0.00)$ & $\mathbf{-0.03}\,(0.00)$ & $\mathbf{-0.03}\,(0.00)$ & $-0.02\,(0.00)$          & $\mathbf{-0.01}\,(0.00)$ & $\mathbf{-0.01}\,(0.00)$ & $-0.04\,(0.01)$          & $-0.04\,(0.01)$          \\
      $q^\star=0.5$ & $-0.03\,(0.00)$           & $\mathbf{-0.02}\,(0.00)$   & $-0.02\,(0.00)$          & $\mathbf{-0.02}\,(0.00)$ & $-0.02\,(0.00)$          & $-0.01\,(0.00)$          & $\mathbf{-0.00}\,(0.00)$ & $\mathbf{-0.00}\,(0.00)$ & $-0.02\,(0.00)$          & $-0.02\,(0.00)$          \\
      $q^\star=0.6$ & $\mathbf{-0.04}\,(0.00)$  & $\mathbf{-0.04}\,(0.01)$   & $-0.05\,(0.01)$          & $\mathbf{-0.04}\,(0.01)$ & $-0.05\,(0.01)$          & $\mathbf{-0.01}\,(0.00)$ & $-0.04\,(0.02)$          & $-0.04\,(0.02)$          & $-0.06\,(0.02)$          & $-0.05\,(0.02)$          \\
      $q^\star=0.7$ & $-0.08\,(0.02)$           & $\mathbf{-0.07}\,(0.00)$   & $-0.07\,(0.00)$          & $\mathbf{-0.07}\,(0.00)$ & $-0.07\,(0.00)$          & $\mathbf{-0.02}\,(0.00)$ & $-0.06\,(0.02)$          & $-0.06\,(0.02)$          & $-0.07\,(0.01)$          & $-0.07\,(0.01)$          \\
      $q^\star=0.8$ & $-0.11\,(0.02)$           & $\mathbf{-0.05}\,(0.00)$   & $-0.05\,(0.01)$          & $\mathbf{-0.05}\,(0.00)$ & $-0.05\,(0.01)$          & $\mathbf{-0.03}\,(0.00)$ & $-0.05\,(0.01)$          & $-0.06\,(0.01)$          & $-0.06\,(0.01)$          & $-0.07\,(0.01)$          \\
      $q^\star=0.9$ & $-0.14\,(0.03)$           & $\mathbf{-0.03}\,(0.00)$   & $\mathbf{-0.03}\,(0.00)$ & $\mathbf{-0.03}\,(0.00)$ & $\mathbf{-0.03}\,(0.00)$ & $\mathbf{-0.03}\,(0.00)$ & $-0.04\,(0.01)$          & $-0.07\,(0.02)$          & $-0.06\,(0.02)$          & $-0.09\,(0.02)$          \\
      $q^\star=1.0$ & $-0.20\,(0.03)$           & $\mathbf{-0.04}\,(0.00)$   & $\mathbf{-0.04}\,(0.00)$ & $\mathbf{-0.04}\,(0.00)$ & $\mathbf{-0.04}\,(0.00)$ & $-0.06\,(0.01)$          & $\mathbf{-0.03}\,(0.02)$ & $\mathbf{-0.03}\,(0.02)$ & $-0.08\,(0.04)$          & $-0.09\,(0.04)$          \\
      \bottomrule
    \end{tabular}}
\end{table}

\begin{table}[h]
  \centering
  \caption{Frequency match, $m=6$. Rank-out-of-64 of the snapped prompt (1 = enumerated optimum), averaged over $n=10$ random seeds; standard error in parentheses. Best method per row in bold (within $0.05$). Prompts with $J$ within $10^{-5}$ are assigned the same rank.}
  \label{tab:freq-match-rank}
  \resizebox{0.85\textwidth}{!}{%
    \begin{tabular}{l rrrrr}
      \toprule
                    & GCG                    & PPT-RB (A)              & PPT-RB                 & PPT (A)                 & PPT                    \\
      \midrule
      \multicolumn{6}{l}{\emph{Beta--Bernoulli ($k=0$)}}                                                                                           \\
      $q^\star=0.0$ & $\mathbf{1.0}\,(0.00)$ & $\mathbf{1.0}\,(0.00)$  & $\mathbf{1.0}\,(0.00)$ & $\mathbf{1.0}\,(0.00)$  & $\mathbf{1.0}\,(0.00)$ \\
      $q^\star=0.1$ & $\mathbf{1.0}\,(0.00)$ & $\mathbf{1.0}\,(0.00)$  & $\mathbf{1.0}\,(0.00)$ & $\mathbf{1.0}\,(0.00)$  & $\mathbf{1.0}\,(0.00)$ \\
      $q^\star=0.2$ & $\mathbf{1.0}\,(0.00)$ & $\mathbf{1.0}\,(0.00)$  & $\mathbf{1.0}\,(0.00)$ & $\mathbf{1.0}\,(0.00)$  & $\mathbf{1.0}\,(0.00)$ \\
      $q^\star=0.3$ & $1.5\,(0.50)$          & $\mathbf{1.0}\,(0.00)$  & $\mathbf{1.0}\,(0.00)$ & $\mathbf{1.0}\,(0.00)$  & $\mathbf{1.0}\,(0.00)$ \\
      $q^\star=0.4$ & $2.0\,(1.00)$          & $\mathbf{1.0}\,(0.00)$  & $\mathbf{1.0}\,(0.00)$ & $\mathbf{1.0}\,(0.00)$  & $\mathbf{1.0}\,(0.00)$ \\
      $q^\star=0.5$ & $\mathbf{1.0}\,(0.00)$ & $\mathbf{1.0}\,(0.00)$  & $\mathbf{1.0}\,(0.00)$ & $\mathbf{1.0}\,(0.00)$  & $\mathbf{1.0}\,(0.00)$ \\
      $q^\star=0.6$ & $\mathbf{4.0}\,(1.53)$ & $11.0\,(0.00)$          & $11.0\,(0.00)$         & $11.0\,(0.00)$          & $11.0\,(0.00)$         \\
      $q^\star=0.7$ & $\mathbf{1.0}\,(0.00)$ & $6.0\,(0.00)$           & $6.0\,(0.00)$          & $6.0\,(0.00)$           & $6.0\,(0.00)$          \\
      $q^\star=0.8$ & $\mathbf{1.0}\,(0.00)$ & $\mathbf{1.0}\,(0.00)$  & $\mathbf{1.0}\,(0.00)$ & $\mathbf{1.0}\,(0.00)$  & $\mathbf{1.0}\,(0.00)$ \\
      $q^\star=0.9$ & $\mathbf{1.0}\,(0.00)$ & $\mathbf{1.0}\,(0.00)$  & $\mathbf{1.0}\,(0.00)$ & $\mathbf{1.0}\,(0.00)$  & $\mathbf{1.0}\,(0.00)$ \\
      $q^\star=1.0$ & $\mathbf{1.0}\,(0.00)$ & $\mathbf{1.0}\,(0.00)$  & $\mathbf{1.0}\,(0.00)$ & $\mathbf{1.0}\,(0.00)$  & $\mathbf{1.0}\,(0.00)$ \\
      \midrule
      \multicolumn{6}{l}{\emph{Reinforced urn ($k=1$)}}                                                                                            \\
      $q^\star=0.0$ & $16.0\,(4.91)$         & $\mathbf{3.9}\,(2.90)$  & $\mathbf{3.9}\,(2.90)$ & $\mathbf{3.9}\,(2.90)$  & $\mathbf{3.9}\,(2.90)$ \\
      $q^\star=0.1$ & $15.3\,(5.08)$         & $\mathbf{3.9}\,(2.90)$  & $\mathbf{3.9}\,(2.90)$ & $\mathbf{3.9}\,(2.90)$  & $\mathbf{3.9}\,(2.90)$ \\
      $q^\star=0.2$ & $15.5\,(5.63)$         & $\mathbf{3.1}\,(2.10)$  & $\mathbf{3.1}\,(2.10)$ & $\mathbf{3.1}\,(2.10)$  & $\mathbf{3.1}\,(2.10)$ \\
      $q^\star=0.3$ & $10.1\,(3.87)$         & $\mathbf{8.3}\,(2.10)$  & $13.4\,(0.93)$         & $\mathbf{8.3}\,(2.10)$  & $13.4\,(0.93)$         \\
      $q^\star=0.4$ & $\mathbf{2.0}\,(0.30)$ & $\mathbf{2.0}\,(0.00)$  & $\mathbf{2.0}\,(0.00)$ & $\mathbf{2.0}\,(0.00)$  & $\mathbf{2.0}\,(0.00)$ \\
      $q^\star=0.5$ & $2.2\,(0.76)$          & $\mathbf{1.4}\,(0.16)$  & $2.0\,(0.00)$          & $1.5\,(0.17)$           & $2.0\,(0.00)$          \\
      $q^\star=0.6$ & $4.7\,(1.14)$          & $\mathbf{4.5}\,(2.50)$  & $7.0\,(3.33)$          & $\mathbf{4.5}\,(2.50)$  & $7.0\,(3.33)$          \\
      $q^\star=0.7$ & $13.2\,(4.73)$         & $\mathbf{11.6}\,(2.37)$ & $14.4\,(1.58)$         & $\mathbf{11.6}\,(2.37)$ & $14.4\,(1.58)$         \\
      $q^\star=0.8$ & $17.7\,(4.52)$         & $\mathbf{1.0}\,(0.00)$  & $3.1\,(2.10)$          & $\mathbf{1.0}\,(0.00)$  & $3.1\,(2.10)$          \\
      $q^\star=0.9$ & $16.7\,(4.11)$         & $\mathbf{1.0}\,(0.00)$  & $\mathbf{1.0}\,(0.00)$ & $\mathbf{1.0}\,(0.00)$  & $\mathbf{1.0}\,(0.00)$ \\
      $q^\star=1.0$ & $16.5\,(4.17)$         & $\mathbf{1.0}\,(0.00)$  & $\mathbf{1.0}\,(0.00)$ & $\mathbf{1.0}\,(0.00)$  & $\mathbf{1.0}\,(0.00)$ \\
      \bottomrule
    \end{tabular}}
\end{table}

\paragraph{Dyck validity.} On Beta--Bernoulli at $m=6$, all four \textsc{PPT} variants reach the enumerated optimum on every seed; \textsc{GCG} averages a worse rank but is within $10^{-4}$ of the optimal $J$. At $m=50$ on Beta--Bernoulli, all methods achieve $J\approx 0.12$. On the reinforced urn at $m=6$ (Tables~\ref{tab:dyck-summary-table} and \ref{tab:dyck-rank}), \textsc{PPT-RB} reaches the enumerated optimum on every seed (rank 1 throughout), \textsc{PPT} is close ($J=0.54$ versus the optimum $J=0.59$), and the remaining methods (\textsc{PPT (analytic)}, \textsc{PPT-RB (analytic)}, \textsc{GCG}) produce $J$ between $0.29$ and $0.38$. At $m=50$ on the reinforced urn the picture inverts: \textsc{GCG} reaches $J=0.61$ while all four \textsc{PPT} variants produce $J$ between $0.014$ and $0.021$. The final $\mathrm{ESS}/L$ for \textsc{PPT-RB} on this configuration is $\approx 0.07$ (mean over 10 seeds), down from $\approx 0.22$ at $m=6$.

\begin{table}[h]
  \centering
  \caption{Dyck validity. Each cell is the mean of $J$ over $n=10$ random seeds (max convention; higher better), with standard error in parentheses. Best method per row in bold (within $10^{-3}$).}
  \label{tab:dyck-summary-table}
  \begin{tabular}{l l rrrrr}
    \toprule
    BFT                     & $m$ & GCG                      & PPT-RB (A)               & PPT-RB                   & PPT (A)                  & PPT                      \\
    \midrule
    Beta--Bernoulli ($k=0$) & 6   & $\mathbf{+0.10}\,(0.00)$ & $\mathbf{+0.10}\,(0.00)$ & $\mathbf{+0.10}\,(0.00)$ & $\mathbf{+0.10}\,(0.00)$ & $\mathbf{+0.10}\,(0.00)$ \\
    Beta--Bernoulli ($k=0$) & 50  & $\mathbf{+0.12}\,(0.00)$ & $\mathbf{+0.12}\,(0.00)$ & $\mathbf{+0.12}\,(0.00)$ & $\mathbf{+0.12}\,(0.00)$ & $\mathbf{+0.12}\,(0.00)$ \\
    Reinforced urn ($k=1$)  & 6   & $+0.29\,(0.05)$          & $+0.33\,(0.09)$          & $\mathbf{+0.59}\,(0.00)$ & $+0.38\,(0.09)$          & $+0.54\,(0.05)$          \\
    Reinforced urn ($k=1$)  & 50  & $\mathbf{+0.61}\,(0.04)$ & $+0.01\,(0.00)$          & $+0.01\,(0.00)$          & $+0.02\,(0.01)$          & $+0.02\,(0.01)$          \\
    \bottomrule
  \end{tabular}
\end{table}

\begin{table}[h]
  \centering
  \caption{Dyck validity, $m=6$. Rank-out-of-64 of the snapped prompt (1 = enumerated optimum), averaged over $n=10$ random seeds; standard error in parentheses. Best method per row in bold (within $0.05$).}
  \label{tab:dyck-rank}
  \begin{tabular}{l rrrrr}
    \toprule
    BFT                     & GCG            & PPT-RB (A)             & PPT-RB                 & PPT (A)                & PPT                    \\
    \midrule
    Beta--Bernoulli ($k=0$) & $3.0\,(1.33)$  & $\mathbf{1.0}\,(0.00)$ & $\mathbf{1.0}\,(0.00)$ & $\mathbf{1.0}\,(0.00)$ & $\mathbf{1.0}\,(0.00)$ \\
    Reinforced urn ($k=1$)  & $12.3\,(4.65)$ & $21.5\,(6.83)$         & $\mathbf{1.0}\,(0.00)$ & $17.4\,(6.70)$         & $4.3\,(3.30)$          \\
    \bottomrule
  \end{tabular}
\end{table}

\FloatBarrier
\subsection{PMC and ESS diagnostics}
\label{app:pmc-ess}

For PPT and PPT-RB, each optimization run logs the importance-weight effective sample size $\mathrm{ESS} := \big(\sum_\ell w_\ell\big)^2 / \sum_\ell w_\ell^2$, with $w_\ell = W(\tilde k^{(\ell)};\,\kappa)$, at every step. Low $\mathrm{ESS}/L$ indicates the tilted posterior $\Pi_\mathrm{tilt}$ has drifted far from the BFT's latent prior, degrading the IS gradient estimate. Tables~\ref{tab:ess-k0} and \ref{tab:ess-k1} report the final $\mathrm{ESS}/L$ (mean, min, max over 10 random seeds) for each (utility, method) configuration at $m=6$; Tables~\ref{tab:ess-k0-m50} and \ref{tab:ess-k1-m50} report the same at $m=50$.

\begin{table}[H]
  \centering
  \caption{Final ESS/$L$ (mean / min / max over 10 seeds), Beta--Bernoulli ($k=0$), $m=6$.}
  \label{tab:ess-k0}
  \resizebox{\textwidth}{!}{%
    \begin{tabular}{lrrrrrrrrrrrr}
      \toprule
      Utility          & \multicolumn{3}{c}{PPT-RB (A)} & \multicolumn{3}{c}{PPT-RB} & \multicolumn{3}{c}{PPT (A)} & \multicolumn{3}{c}{PPT}                                                         \\
                       & mean                           & min                        & max                         & mean                    & min  & max  & mean & min  & max  & mean & min  & max  \\
      \midrule
      Dyck             & 1.00                           & 1.00                       & 1.00                        & 1.00                    & 1.00 & 1.00 & 1.00 & 0.99 & 1.00 & 0.99 & 0.96 & 1.00 \\
      $q^\star=0.0$    & 0.31                           & 0.31                       & 0.32                        & 0.31                    & 0.31 & 0.31 & 0.32 & 0.31 & 0.32 & 0.31 & 0.31 & 0.31 \\
      $q^\star=0.1$    & 0.31                           & 0.31                       & 0.32                        & 0.31                    & 0.31 & 0.31 & 0.32 & 0.31 & 0.33 & 0.31 & 0.31 & 0.32 \\
      $q^\star=0.2$    & 0.31                           & 0.31                       & 0.32                        & 0.31                    & 0.31 & 0.31 & 0.32 & 0.31 & 0.33 & 0.31 & 0.31 & 0.32 \\
      $q^\star=0.3$    & 0.34                           & 0.32                       & 0.36                        & 0.33                    & 0.32 & 0.34 & 0.34 & 0.32 & 0.36 & 0.34 & 0.32 & 0.35 \\
      $q^\star=0.4$    & 0.47                           & 0.46                       & 0.48                        & 0.47                    & 0.46 & 0.47 & 0.47 & 0.46 & 0.48 & 0.47 & 0.46 & 0.47 \\
      $q^\star=0.5$    & 1.00                           & 1.00                       & 1.00                        & 1.00                    & 1.00 & 1.00 & 1.00 & 0.99 & 1.00 & 1.00 & 1.00 & 1.00 \\
      $q^\star=0.6$    & 0.48                           & 0.47                       & 0.52                        & 0.49                    & 0.48 & 0.50 & 0.48 & 0.46 & 0.49 & 0.48 & 0.48 & 0.49 \\
      $q^\star=0.7$    & 0.34                           & 0.32                       & 0.36                        & 0.35                    & 0.33 & 0.36 & 0.34 & 0.32 & 0.36 & 0.35 & 0.33 & 0.36 \\
      $q^\star=0.8$    & 0.32                           & 0.31                       & 0.33                        & 0.32                    & 0.32 & 0.32 & 0.32 & 0.32 & 0.34 & 0.32 & 0.32 & 0.33 \\
      $q^\star=0.9$    & 0.32                           & 0.31                       & 0.33                        & 0.32                    & 0.32 & 0.32 & 0.32 & 0.31 & 0.34 & 0.32 & 0.32 & 0.33 \\
      $q^\star=1.0$    & 0.32                           & 0.31                       & 0.33                        & 0.32                    & 0.32 & 0.32 & 0.32 & 0.31 & 0.33 & 0.32 & 0.32 & 0.33 \\
      $\tau^\star=0.1$ & 0.31                           & 0.31                       & 0.32                        & 0.31                    & 0.31 & 0.31 & 0.32 & 0.31 & 0.32 & 0.31 & 0.31 & 0.31 \\
      $\tau^\star=0.2$ & 0.31                           & 0.31                       & 0.32                        & 0.31                    & 0.31 & 0.31 & 0.32 & 0.31 & 0.32 & 0.31 & 0.31 & 0.31 \\
      $\tau^\star=0.3$ & 0.31                           & 0.31                       & 0.32                        & 0.31                    & 0.31 & 0.31 & 0.32 & 0.31 & 0.32 & 0.31 & 0.31 & 0.31 \\
      $\tau^\star=0.4$ & 0.31                           & 0.31                       & 0.32                        & 0.31                    & 0.31 & 0.31 & 0.32 & 0.31 & 0.32 & 0.31 & 0.31 & 0.31 \\
      $\tau^\star=0.6$ & 0.32                           & 0.31                       & 0.33                        & 0.32                    & 0.32 & 0.32 & 0.32 & 0.31 & 0.33 & 0.32 & 0.32 & 0.32 \\
      $\tau^\star=0.7$ & 0.32                           & 0.31                       & 0.33                        & 0.32                    & 0.32 & 0.32 & 0.32 & 0.31 & 0.33 & 0.32 & 0.32 & 0.32 \\
      $\tau^\star=0.8$ & 0.32                           & 0.31                       & 0.33                        & 0.32                    & 0.32 & 0.32 & 0.32 & 0.31 & 0.33 & 0.32 & 0.32 & 0.32 \\
      $\tau^\star=0.9$ & 0.32                           & 0.31                       & 0.33                        & 0.32                    & 0.32 & 0.32 & 0.32 & 0.31 & 0.33 & 0.32 & 0.32 & 0.32 \\
      \bottomrule
    \end{tabular}}
\end{table}

\begin{table}[H]
  \centering
  \caption{Final ESS/$L$ (mean / min / max over 10 seeds), Reinforced urn ($k=1$), $m=6$.}
  \label{tab:ess-k1}
  \resizebox{\textwidth}{!}{%
    \begin{tabular}{lrrrrrrrrrrrr}
      \toprule
      Utility       & \multicolumn{3}{c}{PPT-RB (A)} & \multicolumn{3}{c}{PPT-RB} & \multicolumn{3}{c}{PPT (A)} & \multicolumn{3}{c}{PPT}                                                         \\
                    & mean                           & min                        & max                         & mean                    & min  & max  & mean & min  & max  & mean & min  & max  \\
      \midrule
      Dyck          & 0.22                           & 0.21                       & 0.23                        & 0.20                    & 0.20 & 0.20 & 0.22 & 0.21 & 0.23 & 0.22 & 0.20 & 0.42 \\
      $q^\star=0.0$ & 0.31                           & 0.22                       & 0.35                        & 0.41                    & 0.20 & 0.49 & 0.31 & 0.22 & 0.35 & 0.42 & 0.20 & 0.49 \\
      $q^\star=0.1$ & 0.31                           & 0.22                       & 0.35                        & 0.41                    & 0.20 & 0.49 & 0.32 & 0.22 & 0.35 & 0.42 & 0.20 & 0.49 \\
      $q^\star=0.2$ & 0.31                           & 0.22                       & 0.35                        & 0.41                    & 0.20 & 0.49 & 0.32 & 0.22 & 0.35 & 0.42 & 0.20 & 0.49 \\
      $q^\star=0.3$ & 0.26                           & 0.22                       & 0.30                        & 0.25                    & 0.20 & 0.44 & 0.27 & 0.22 & 0.31 & 0.25 & 0.20 & 0.44 \\
      $q^\star=0.4$ & 0.22                           & 0.22                       & 0.23                        & 0.20                    & 0.20 & 0.20 & 0.23 & 0.22 & 0.24 & 0.20 & 0.20 & 0.21 \\
      $q^\star=0.5$ & 0.22                           & 0.21                       & 0.23                        & 0.20                    & 0.20 & 0.20 & 0.22 & 0.21 & 0.23 & 0.20 & 0.20 & 0.20 \\
      $q^\star=0.6$ & 0.24                           & 0.22                       & 0.35                        & 0.24                    & 0.20 & 0.38 & 0.24 & 0.22 & 0.35 & 0.24 & 0.20 & 0.38 \\
      $q^\star=0.7$ & 0.28                           & 0.22                       & 0.35                        & 0.25                    & 0.20 & 0.38 & 0.29 & 0.23 & 0.35 & 0.25 & 0.20 & 0.38 \\
      $q^\star=0.8$ & 0.32                           & 0.27                       & 0.35                        & 0.32                    & 0.20 & 0.38 & 0.32 & 0.27 & 0.35 & 0.32 & 0.20 & 0.38 \\
      $q^\star=0.9$ & 0.32                           & 0.27                       & 0.35                        & 0.34                    & 0.25 & 0.38 & 0.32 & 0.27 & 0.35 & 0.34 & 0.25 & 0.38 \\
      $q^\star=1.0$ & 0.32                           & 0.27                       & 0.35                        & 0.34                    & 0.25 & 0.38 & 0.32 & 0.27 & 0.35 & 0.34 & 0.25 & 0.38 \\
      sym-0.1       & 0.22                           & 0.21                       & 0.23                        & 0.20                    & 0.20 & 0.20 & 0.22 & 0.21 & 0.23 & 0.20 & 0.20 & 0.20 \\
      sym-0.2       & 0.22                           & 0.21                       & 0.23                        & 0.20                    & 0.20 & 0.20 & 0.22 & 0.21 & 0.23 & 0.20 & 0.20 & 0.20 \\
      sym-0.3       & 0.22                           & 0.21                       & 0.23                        & 0.20                    & 0.20 & 0.20 & 0.22 & 0.21 & 0.23 & 0.20 & 0.20 & 0.20 \\
      sym-0.4       & 0.22                           & 0.21                       & 0.23                        & 0.20                    & 0.20 & 0.20 & 0.22 & 0.21 & 0.23 & 0.20 & 0.20 & 0.20 \\
      sym-0.6       & 0.34                           & 0.28                       & 0.35                        & 0.41                    & 0.38 & 0.48 & 0.33 & 0.28 & 0.36 & 0.38 & 0.26 & 0.49 \\
      sym-0.7       & 0.34                           & 0.27                       & 0.35                        & 0.41                    & 0.38 & 0.48 & 0.33 & 0.28 & 0.36 & 0.38 & 0.26 & 0.49 \\
      sym-0.8       & 0.34                           & 0.27                       & 0.35                        & 0.41                    & 0.38 & 0.48 & 0.33 & 0.28 & 0.36 & 0.38 & 0.26 & 0.49 \\
      sym-0.9       & 0.34                           & 0.27                       & 0.35                        & 0.41                    & 0.38 & 0.48 & 0.33 & 0.28 & 0.36 & 0.38 & 0.26 & 0.48 \\
      sym-1.0       & 0.34                           & 0.33                       & 0.35                        & 0.44                    & 0.38 & 0.48 & 0.34 & 0.33 & 0.35 & 0.44 & 0.38 & 0.48 \\
      dir-0         & 0.32                           & 0.27                       & 0.35                        & 0.32                    & 0.22 & 0.38 & 0.32 & 0.27 & 0.35 & 0.32 & 0.21 & 0.38 \\
      dir-1         & 0.33                           & 0.29                       & 0.35                        & 0.44                    & 0.37 & 0.49 & 0.33 & 0.29 & 0.35 & 0.44 & 0.37 & 0.49 \\
      dir-2         & 0.32                           & 0.27                       & 0.35                        & 0.34                    & 0.24 & 0.38 & 0.32 & 0.27 & 0.35 & 0.34 & 0.24 & 0.38 \\
      dir-3         & 0.34                           & 0.27                       & 0.35                        & 0.40                    & 0.25 & 0.48 & 0.33 & 0.27 & 0.36 & 0.38 & 0.25 & 0.49 \\
      dir-4         & 0.34                           & 0.30                       & 0.35                        & 0.43                    & 0.38 & 0.48 & 0.34 & 0.30 & 0.35 & 0.43 & 0.38 & 0.49 \\
      dir-5         & 0.22                           & 0.21                       & 0.23                        & 0.20                    & 0.20 & 0.20 & 0.22 & 0.21 & 0.23 & 0.20 & 0.20 & 0.20 \\
      dir-6         & 0.34                           & 0.31                       & 0.35                        & 0.45                    & 0.38 & 0.48 & 0.34 & 0.31 & 0.35 & 0.43 & 0.38 & 0.49 \\
      dir-7         & 0.34                           & 0.33                       & 0.35                        & 0.42                    & 0.38 & 0.48 & 0.33 & 0.28 & 0.36 & 0.39 & 0.27 & 0.48 \\
      dir-8         & 0.25                           & 0.22                       & 0.35                        & 0.24                    & 0.20 & 0.38 & 0.25 & 0.22 & 0.35 & 0.24 & 0.20 & 0.38 \\
      dir-9         & 0.33                           & 0.29                       & 0.35                        & 0.41                    & 0.20 & 0.49 & 0.33 & 0.29 & 0.35 & 0.41 & 0.20 & 0.49 \\
      \bottomrule
    \end{tabular}}
\end{table}

\begin{table}[H]
  \centering
  \caption{Final ESS/$L$ (mean / min / max over 10 seeds), Beta--Bernoulli ($k=0$), $m=50$.}
  \label{tab:ess-k0-m50}
  \resizebox{\textwidth}{!}{%
    \begin{tabular}{lrrrrrrrrrrrr}
      \toprule
      Utility          & \multicolumn{3}{c}{PPT-RB (A)} & \multicolumn{3}{c}{PPT-RB} & \multicolumn{3}{c}{PPT (A)} & \multicolumn{3}{c}{PPT}                                                         \\
                       & mean                           & min                        & max                         & mean                    & min  & max  & mean & min  & max  & mean & min  & max  \\
      \midrule
      Dyck             & 1.00                           & 1.00                       & 1.00                        & 1.00                    & 1.00 & 1.00 & 0.97 & 0.84 & 1.00 & 0.94 & 0.86 & 1.00 \\
      $q^\star=0.0$    & 0.11                           & 0.10                       & 0.12                        & 0.11                    & 0.11 & 0.11 & 0.12 & 0.10 & 0.13 & 0.12 & 0.11 & 0.13 \\
      $q^\star=0.1$    & 0.12                           & 0.10                       & 0.13                        & 0.11                    & 0.11 & 0.12 & 0.12 & 0.10 & 0.13 & 0.12 & 0.11 & 0.13 \\
      $q^\star=0.2$    & 0.16                           & 0.13                       & 0.21                        & 0.16                    & 0.12 & 0.21 & 0.16 & 0.13 & 0.21 & 0.16 & 0.12 & 0.22 \\
      $q^\star=0.3$    & 0.20                           & 0.13                       & 0.38                        & 0.17                    & 0.13 & 0.31 & 0.19 & 0.13 & 0.31 & 0.18 & 0.13 & 0.32 \\
      $q^\star=0.4$    & 0.40                           & 0.21                       & 0.48                        & 0.42                    & 0.24 & 0.47 & 0.43 & 0.23 & 0.50 & 0.42 & 0.23 & 0.49 \\
      $q^\star=0.5$    & 1.00                           & 1.00                       & 1.00                        & 1.00                    & 1.00 & 1.00 & 0.99 & 0.96 & 1.00 & 0.99 & 0.97 & 1.00 \\
      $q^\star=0.6$    & 0.36                           & 0.15                       & 0.49                        & 0.37                    & 0.16 & 0.48 & 0.36 & 0.16 & 0.50 & 0.37 & 0.16 & 0.51 \\
      $q^\star=0.7$    & 0.19                           & 0.12                       & 0.33                        & 0.20                    & 0.13 & 0.33 & 0.19 & 0.12 & 0.32 & 0.18 & 0.13 & 0.33 \\
      $q^\star=0.8$    & 0.17                           & 0.12                       & 0.22                        & 0.17                    & 0.13 & 0.21 & 0.17 & 0.12 & 0.23 & 0.17 & 0.13 & 0.21 \\
      $q^\star=0.9$    & 0.12                           & 0.11                       & 0.13                        & 0.12                    & 0.11 & 0.13 & 0.12 & 0.11 & 0.14 & 0.12 & 0.11 & 0.13 \\
      $q^\star=1.0$    & 0.11                           & 0.11                       & 0.12                        & 0.11                    & 0.11 & 0.12 & 0.12 & 0.11 & 0.14 & 0.12 & 0.11 & 0.13 \\
      $\tau^\star=0.1$ & 0.11                           & 0.10                       & 0.11                        & 0.11                    & 0.11 & 0.11 & 0.12 & 0.10 & 0.13 & 0.12 & 0.11 & 0.12 \\
      $\tau^\star=0.2$ & 0.11                           & 0.10                       & 0.11                        & 0.11                    & 0.11 & 0.11 & 0.12 & 0.10 & 0.13 & 0.12 & 0.11 & 0.12 \\
      $\tau^\star=0.3$ & 0.11                           & 0.10                       & 0.11                        & 0.11                    & 0.11 & 0.11 & 0.12 & 0.10 & 0.13 & 0.12 & 0.11 & 0.12 \\
      $\tau^\star=0.4$ & 0.11                           & 0.10                       & 0.11                        & 0.11                    & 0.11 & 0.11 & 0.12 & 0.10 & 0.13 & 0.12 & 0.11 & 0.12 \\
      $\tau^\star=0.6$ & 0.11                           & 0.11                       & 0.12                        & 0.11                    & 0.11 & 0.11 & 0.12 & 0.11 & 0.13 & 0.12 & 0.11 & 0.13 \\
      $\tau^\star=0.7$ & 0.11                           & 0.11                       & 0.12                        & 0.11                    & 0.11 & 0.11 & 0.12 & 0.11 & 0.13 & 0.12 & 0.11 & 0.13 \\
      $\tau^\star=0.8$ & 0.11                           & 0.11                       & 0.12                        & 0.11                    & 0.11 & 0.11 & 0.12 & 0.11 & 0.13 & 0.12 & 0.11 & 0.13 \\
      $\tau^\star=0.9$ & 0.11                           & 0.11                       & 0.12                        & 0.11                    & 0.11 & 0.11 & 0.12 & 0.11 & 0.13 & 0.12 & 0.11 & 0.13 \\
      \bottomrule
    \end{tabular}}
\end{table}

\begin{table}[H]
  \centering
  \caption{Final ESS/$L$ (mean / min / max over 10 seeds), Reinforced urn ($k=1$), $m=50$.}
  \label{tab:ess-k1-m50}
  \resizebox{\textwidth}{!}{%
    \begin{tabular}{lrrrrrrrrrrrr}
      \toprule
      Utility       & \multicolumn{3}{c}{PPT-RB (A)} & \multicolumn{3}{c}{PPT-RB} & \multicolumn{3}{c}{PPT (A)} & \multicolumn{3}{c}{PPT}                                                         \\
                    & mean                           & min                        & max                         & mean                    & min  & max  & mean & min  & max  & mean & min  & max  \\
      \midrule
      Dyck          & 0.05                           & 0.04                       & 0.09                        & 0.07                    & 0.06 & 0.12 & 0.06 & 0.04 & 0.16 & 0.08 & 0.06 & 0.20 \\
      $q^\star=0.0$ & 0.11                           & 0.09                       & 0.12                        & 0.15                    & 0.11 & 0.26 & 0.09 & 0.03 & 0.12 & 0.12 & 0.04 & 0.26 \\
      $q^\star=0.1$ & 0.13                           & 0.09                       & 0.18                        & 0.19                    & 0.12 & 0.29 & 0.10 & 0.03 & 0.18 & 0.14 & 0.04 & 0.29 \\
      $q^\star=0.2$ & 0.18                           & 0.03                       & 0.30                        & 0.21                    & 0.03 & 0.36 & 0.15 & 0.03 & 0.32 & 0.17 & 0.04 & 0.36 \\
      $q^\star=0.3$ & 0.07                           & 0.02                       & 0.11                        & 0.08                    & 0.03 & 0.15 & 0.07 & 0.03 & 0.11 & 0.09 & 0.04 & 0.14 \\
      $q^\star=0.4$ & 0.03                           & 0.02                       & 0.03                        & 0.03                    & 0.03 & 0.03 & 0.04 & 0.03 & 0.10 & 0.06 & 0.04 & 0.12 \\
      $q^\star=0.5$ & 0.03                           & 0.02                       & 0.03                        & 0.03                    & 0.03 & 0.03 & 0.04 & 0.03 & 0.04 & 0.05 & 0.04 & 0.05 \\
      $q^\star=0.6$ & 0.04                           & 0.02                       & 0.12                        & 0.05                    & 0.03 & 0.14 & 0.05 & 0.03 & 0.12 & 0.06 & 0.04 & 0.14 \\
      $q^\star=0.7$ & 0.07                           & 0.03                       & 0.12                        & 0.07                    & 0.04 & 0.14 & 0.07 & 0.03 & 0.12 & 0.07 & 0.04 & 0.14 \\
      $q^\star=0.8$ & 0.17                           & 0.03                       & 0.28                        & 0.12                    & 0.04 & 0.26 & 0.13 & 0.03 & 0.27 & 0.12 & 0.04 & 0.26 \\
      $q^\star=0.9$ & 0.11                           & 0.03                       & 0.18                        & 0.11                    & 0.03 & 0.21 & 0.11 & 0.03 & 0.18 & 0.11 & 0.03 & 0.21 \\
      $q^\star=1.0$ & 0.10                           & 0.03                       & 0.12                        & 0.11                    & 0.03 & 0.14 & 0.08 & 0.03 & 0.12 & 0.10 & 0.03 & 0.16 \\
      sym-0.1       & 0.03                           & 0.02                       & 0.03                        & 0.03                    & 0.03 & 0.03 & 0.03 & 0.02 & 0.03 & 0.03 & 0.03 & 0.03 \\
      sym-0.2       & 0.03                           & 0.02                       & 0.03                        & 0.03                    & 0.03 & 0.03 & 0.03 & 0.02 & 0.03 & 0.03 & 0.03 & 0.03 \\
      sym-0.3       & 0.03                           & 0.02                       & 0.03                        & 0.03                    & 0.03 & 0.03 & 0.03 & 0.02 & 0.03 & 0.03 & 0.03 & 0.03 \\
      sym-0.4       & 0.03                           & 0.02                       & 0.03                        & 0.05                    & 0.03 & 0.23 & 0.03 & 0.02 & 0.03 & 0.03 & 0.03 & 0.03 \\
      sym-0.6       & 0.11                           & 0.09                       & 0.12                        & 0.13                    & 0.09 & 0.15 & 0.11 & 0.09 & 0.12 & 0.13 & 0.10 & 0.15 \\
      sym-0.7       & 0.11                           & 0.10                       & 0.12                        & 0.14                    & 0.12 & 0.16 & 0.11 & 0.09 & 0.12 & 0.13 & 0.09 & 0.15 \\
      sym-0.8       & 0.11                           & 0.10                       & 0.12                        & 0.14                    & 0.12 & 0.16 & 0.11 & 0.09 & 0.12 & 0.13 & 0.10 & 0.16 \\
      sym-0.9       & 0.11                           & 0.10                       & 0.12                        & 0.14                    & 0.14 & 0.16 & 0.11 & 0.09 & 0.12 & 0.13 & 0.10 & 0.16 \\
      sym-1.0       & 0.11                           & 0.11                       & 0.12                        & 0.14                    & 0.14 & 0.16 & 0.11 & 0.10 & 0.12 & 0.14 & 0.11 & 0.16 \\
      dir-0         & 0.09                           & 0.03                       & 0.12                        & 0.09                    & 0.03 & 0.14 & 0.08 & 0.03 & 0.12 & 0.08 & 0.03 & 0.14 \\
      dir-1         & 0.10                           & 0.08                       & 0.12                        & 0.13                    & 0.03 & 0.15 & 0.09 & 0.02 & 0.12 & 0.12 & 0.03 & 0.15 \\
      dir-2         & 0.11                           & 0.09                       & 0.12                        & 0.12                    & 0.08 & 0.14 & 0.11 & 0.09 & 0.12 & 0.13 & 0.10 & 0.15 \\
      dir-3         & 0.11                           & 0.10                       & 0.12                        & 0.14                    & 0.12 & 0.16 & 0.11 & 0.09 & 0.12 & 0.13 & 0.10 & 0.15 \\
      dir-4         & 0.11                           & 0.09                       & 0.12                        & 0.14                    & 0.11 & 0.16 & 0.11 & 0.09 & 0.12 & 0.14 & 0.11 & 0.16 \\
      dir-5         & 0.03                           & 0.02                       & 0.03                        & 0.03                    & 0.03 & 0.03 & 0.03 & 0.02 & 0.03 & 0.03 & 0.03 & 0.03 \\
      dir-6         & 0.11                           & 0.09                       & 0.12                        & 0.14                    & 0.11 & 0.16 & 0.11 & 0.09 & 0.12 & 0.13 & 0.11 & 0.16 \\
      dir-7         & 0.11                           & 0.11                       & 0.12                        & 0.14                    & 0.12 & 0.16 & 0.11 & 0.09 & 0.12 & 0.14 & 0.10 & 0.16 \\
      dir-8         & 0.05                           & 0.02                       & 0.12                        & 0.06                    & 0.03 & 0.14 & 0.05 & 0.02 & 0.12 & 0.06 & 0.03 & 0.14 \\
      dir-9         & 0.10                           & 0.02                       & 0.12                        & 0.13                    & 0.03 & 0.15 & 0.09 & 0.02 & 0.12 & 0.11 & 0.03 & 0.15 \\
      \bottomrule
    \end{tabular}}
\end{table}

\section{Broader impacts}
\label{app:broader_impacts}

Elicitation methods are dual-use: they support safety testing but could inform adversarial prompt search. Our method is restricted to stylized BFTs with binary vocabularies, limiting near-term risk.

\end{document}